\documentclass[10pt,journal,compsoc]{IEEEtran}
\usepackage[numbers,sort&compress,square]{natbib}
\usepackage{color}

\usepackage{bbm}

\usepackage{subfigure}

\ifCLASSINFOpdf
   \usepackage[pdftex]{graphicx}
\else
\fi
\usepackage{hyperref}

\usepackage{algorithmic}
\usepackage{algorithm}

\usepackage{amsmath}
\usepackage{amsfonts}
\usepackage{amsthm}

\usepackage{multirow}

\usepackage[table]{xcolor}
\usepackage{collcell}
\usepackage{hhline}
\usepackage{pgf}

\newcommand\gray{gray}
    
\newcommand\ColCell[1]{%
  \pgfmathparse{#1<40?1:0}%
    \ifnum\pgfmathresult=0\relax\color{white}\fi
  \pgfmathparse{#1<1?1:0}%
    \ifnum\pgfmathresult=1\relax\color{white}\fi
  \pgfmathparse{1-#1/100}%
  \expandafter\cellcolor\expandafter[%
    \expandafter\gray\expandafter]\expandafter{\pgfmathresult}#1}

\newcolumntype{E}{>{\collectcell\ColCell}c<{\endcollectcell}}

\newtheorem{theorem}{Theorem}
\newtheorem{corollary}{Corollary}[theorem]
\newtheorem{lemma}[theorem]{Lemma}

\newcommand{\revision}[1]{\textcolor{black}{#1}}

\begin{document}
%
\title{Latent Class-Conditional Noise Model}
%
%
%
%

\author{Jiangchao~Yao,
        Bo~Han,
        Zhihan~Zhou,
        Ya~Zhang$^\dagger$,
        and~Ivor~W.~Tsang, \emph{Fellow, IEEE}
\IEEEcompsocitemizethanks{
\IEEEcompsocthanksitem Jiangchao Yao, Zhihan Zhou and Ya Zhang are with Cooperative Medianet Innovation Center,  Shanghai Jiao Tong University, Shanghai 200240, China. Jiangchao Yao and Ya Zhang are also with Shanghai AI Laboratory, Shanghai 200030, China. $\dagger$ means Ya Zhang is the corresponding author. \protect\\
E-mail: \{Sunarker, zhihanzhou, ya\_zhang\}@sjtu.edu.cn
\IEEEcompsocthanksitem Bo Han is with the Department of Computer Science,  Hong Kong Baptist University, Kowloon Tong, Kowloon, Hong Kong, China.\protect\\
E-mail: bhanml@comp.hkbu.edu.hk
\IEEEcompsocthanksitem Ivor W. Tsang is with the A*STAR Centre for Frontier AI Research,
Singapore 138632.\protect\\
E-mail: ivor\_tsang@ihpc.a-star.edu.sg}
}

%
%

\markboth{Journal of \LaTeX\ Class Files,~Vol.~14, No.~8, August~2015}%
{Shell \MakeLowercase{\textit{et al.}}: Bare Demo of IEEEtran.cls for Computer Society Journals}
%



\IEEEtitleabstractindextext{%
\begin{abstract}
Learning with noisy labels has become imperative in the Big Data era, which saves expensive human labors on accurate annotations. Previous noise-transition-based methods have achieved \revision{theoretically}-grounded performance under the Class-Conditional Noise model (CCN). However, these approaches builds upon an ideal but impractical \emph{anchor set} available to pre-estimate the noise transition. Even though subsequent works adapt the estimation as a neural layer, the ill-posed stochastic learning of its parameters in back-propagation easily \revision{falls into undesired} local minimums. We solve this problem by introducing a Latent Class-Conditional Noise model (LCCN) to
parameterize the noise transition under a Bayesian framework. By projecting the noise transition into the Dirichlet space, the learning is constrained on a simplex characterized by the complete dataset, instead of some ad-hoc parametric space wrapped by the neural layer. We then deduce a dynamic label regression method for LCCN, whose Gibbs sampler allows us efficiently infer the latent true labels to train the classifier and to model the noise. Our approach safeguards the stable update of the noise transition, which avoids previous arbitrarily tuning from a mini-batch of samples. We further generalize LCCN to different counterparts compatible \revision{with open-set noisy labels}, \revision{semi-supervised learning as well as cross-model training}. A range of experiments demonstrate the advantages of LCCN and its variants over the current state-of-the-art methods. The code is available at \href{https://github.com/Sunarker/LCCN}{here}.
\end{abstract}

}

\maketitle

\IEEEdisplaynontitleabstractindextext

%
\IEEEpeerreviewmaketitle


\IEEEraisesectionheading{\section{Introduction}\label{sec:introduction}}

%
%
%
%
\IEEEPARstart{L}{arge-scale} datasets with accurate labels 
have driven the success of deep neural networks (DNNs) in computer vision~\cite{NIPS2012_4824}, natural language processing~\cite{sutskever2014sequence}, and speech recognition~\cite{hinton2012deep}. However, for many real-world applications, it is expensive to collect precisely annotated data in large volume. Instead, samples with noisy supervision, as an alternative to alleviate the annotation burden, can be acquired inexhaustibly on social websites and have shown potential to many applications in the deep learning area~\cite{7953515,8293849,8410611}. 

\begin{figure*} 
\centering
\includegraphics[width=0.98\textwidth]{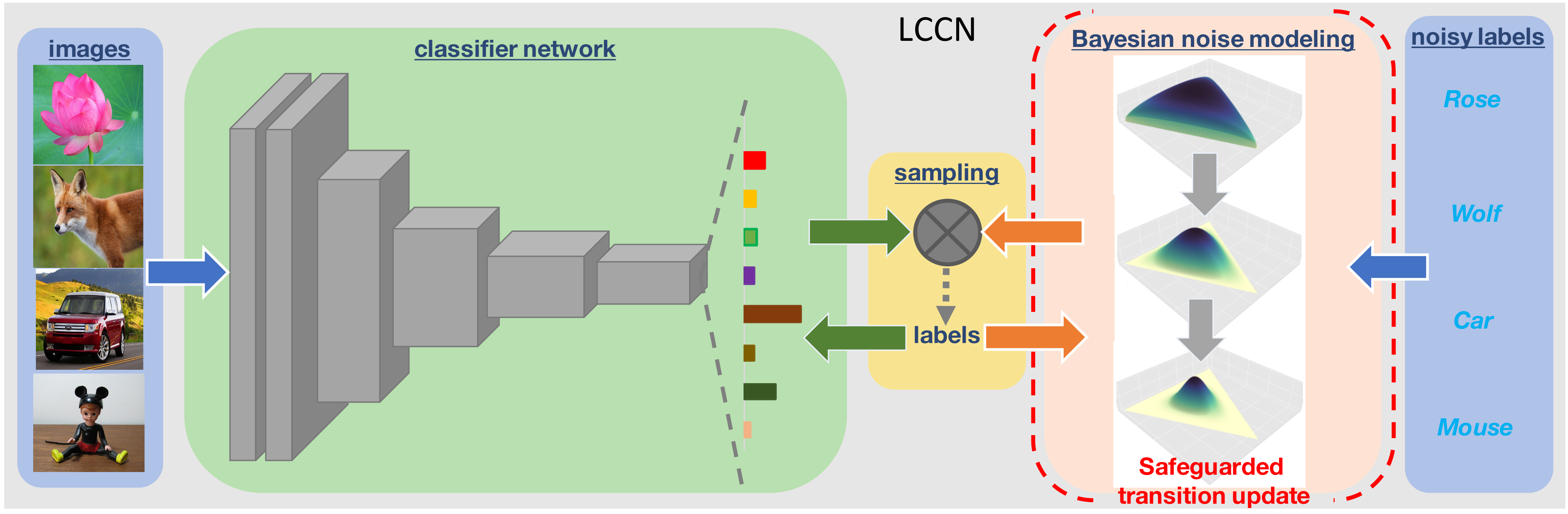}
\caption{Dynamic label regression for LCCN. The images and noisy labels are respectively inputted to the classifier and the safeguarded Bayesian noise modeling to compute the prediction and the conditional transition. Then, the latent labels are sampled based on their product, and then used for the training of the classifier and the safeguarded Bayesian noise modeling. \revision{All components are trained} end-to-end in a stochastic fashion.}\label{fig:network}
\end{figure*}

However, it is challenging to train DNNs in the presence of noisy supervision due to the memorization effect~\cite{arpit2017closer}. To prevent the degeneration under this setting, several seminal works~\cite{han2020survey,yao2018deep} have been explored from the perspective of regularization to avoid overfitting or sample/label refinement. For example, Arpit \textit{et al.}~\cite{arpit2017closer} applied the dropout regularization to decelerate noise memorization, and Reed \textit{et al.}~\cite{reed2014training} explored the self-weighting mechanism via its predictions to weaken the noise effect. 
Different from the aforementioned methodologies,
this study focuses on the third prosperous statistical perspective, learning via noise transition. This line of works place a transition matrix on top of the classifier to disentangle the noise from the learning procedure. Previous works~\cite{liu2016classification,patrini2017making,dualT} presents a two-step solution: first estimate the noise transition, and then freeze it to train the classifier. However, it is usually impractical to have or approximately have an accurate anchor set~\cite{yao2019safeguarded} as the sufficient statistics of the noise transition. The alternative ways~\cite{goldberger2016training,xia2019t_revision} cast the noise transition as a neural layer and adapt it along with the training. Unfortunately, the current stochastic training as a neural layer to noise transition is ill-posed, yielding that the optimization depends on the empirically tuning to dodge the undesired local minimums.
Actually, the reason of this phenomenon is the backpropagation applied in the neural layer could induce the arbitrary update to the noise transition by a mini-batch of samples. The worst case is that some atypical mini-batches containing the extreme noise could catastrophically destroy the well estimation regarding the noise transition parameters for the complete dataset.

To solve this problem, we extend the conventional CCN into a latent counterpart, \textit{i.e.,} Latent Class-Conditional Noise model (LCCN). The intuition is that LCCN that parameterizes the noise transition under a Bayesian framework \revision{can explicitly characterize} the dependency on the complete dataset instead of the mini-batch of sample statistics. This fundamentally constrains the learning of the noise transition unlike the backpropagation applied in the neural layer, yielding a more stable estimation about the noise transition.
Although it apparently builds a more sophisticated model, we show that the optimization for LCCN deduced by Gibbs sampling can be as efficient as stochastic gradient descent. Specifically, we propose a novel dynamic label regression method and illustrate its learning procedure in Fig.~\ref{fig:network} (see the caption for more details). Note that, although our optimization method iteratively infers the latent true labels and applies them for the training of the classifier and the noise modeling, only a small amount of extra computational cost is introduced. We theoretically provide the convergence analysis and the generalization bound about LCCN under the label noise, and prove that our optimization approach safeguards the update of noise transition by a mini-batch.

Beyond the aforementioned property of LCCN, we show that it is easy to extend LCCN under more broader settings. Concretely, by taking the out-of-distribution (OOD) samples into account~\cite{liang2017enhancing}, we increase the dimension of the latent label in LCCN to handle the open-set label noise. With a slight modification in the inference process of the latent true labels, the semi-supervised learning paradigm is seamlessly integrated into the optimization. Considering the advantages of cross-model training like co-teaching and Dividemix~\cite{Han2018CoteachingRT,li2019dividemix}, we use two LCCNs to construct a divideLCCN model, which achieves the state-of-the-art performance on a range of datasets. In summary, the contribution of this paper can be categorized as follows.
\begin{itemize}
\item We propose a Latent Class-Conditional Noise model to explicitly constrain the learning of the noise transition when jointly trained with deep neural networks. Unlike the backpropagation that arbitrarily updates the parameters of the transition layer, we introduce an efficient dynamic label regression\footnote{Note that, we use the word ``regression" to indicate the noisy label is progressively corrected to the groundtruth in the optimization.} to maintain the stable optimization in learning with noisy labels.
\item We theoretically show the convergence property of LCCN by means of the dynamic label regression, and characterize the generalization bound of LCCN to uncover the factors for learning with label noise. Simultaneously, we prove that our optimization of the noise transition via a mini-batch of samples is safely bounded compared to the backpropagation.
\item We show how to extend the original LCCN without much effort to several variants, \textit{i.e.,} LCCN$^*$ that handles open-set noisy labels when considering the OOD samples, and LCCN+ that is compatible with the semi-supervised learning setting when some precisely annotated samples are available,
and divideLCCN that leverages the advantage of the cross-model training~\cite{Han2018CoteachingRT} to boost the performance.
\end{itemize}
We conduct a range of experiments in the popular CIFAR-10, CIFAR-100 datasets and large real-world noisy datasets, Clothing1M and WebVision17. Comprehensive results have demonstrated the superior performance of our model compared with existing state-of-the-art methods. 

The rest part of this paper is organized as follows. Section II briefly reviews the related research of learning with noisy labels in deep learning. Then, we introduce our Latent Class-Conditional model and the dynamic label regression method in Section III, where the corresponding theoretical analysis and the further extension of LCCN is also included. We validate the efficiency of our method over a range of experiments in Section IV. Section V concludes the paper.

\section{Related Work}
Recently, a large amount of methods have been developed for learning with noisy labels in the context of deep learning. There are many perspectives from \emph{data}, \emph{objective} and \emph{optimization} that could summarize these works~\cite{han2020survey}. To simplify the narration of the motivation to our study, we review these works according to the taxonomy of noise transition, sample re-weighting, model regularization, \revision{loss functions in this section. Besides, the EM optimization in label-noise learning that tightly correlates with our motivation is also reviewed.}

\subsection{Learning via Noise Transition}
This branch of research models a noise transition on top of the classifier to minimize the influence of label noise. For example, Sukhbaatar~\textit{et al.,}~\cite{sukhbaatar2014training} introduced a noise transition matrix on top of CNN to learn with noisy supervision. By regularizing the training process, they progressively squeezed the label noise from the base model to the noise matrix.
Misra \textit{et al.,}~\cite{MisraNoisy16} considered the ``reporting bias'' in human-centric labels via marginalization of conditional estimates, which is a special case of noise transition. Patrini \textit{et al.,}~\cite{patrini2017making} theoretically demonstrated: the backward correction with the inverse of the noise transition is unbiased to train the classifier in the presence of noisy labels; the forward noise transition make the training share the same minimizer with that on the clean data. However, the performance critically depends on the accuracy of the pre-estimated noise transition. 
Subsequent improvement in~\cite{goldberger2016training} models the noise transition as a Softmax layer and jointly trains its parameters along with DNNs by the elaborate finetuning. Han \textit{et al.,}~\cite{Han2018MaskingAN} further introduced the prior structure knowledge to constrain the optimization of the noise transition, which avoids some sub-optimal solutions in the hypothesis space. \revision{Xia \textit{et al.,}~\cite{xia2019t_revision} and Yao \textit{et al.,}~\cite{yao2020dual} respectively proposed T-revision  and Dual T method to account for the improvement on the basis of the ordinary pre-estimation. Zhu \textit{et al.,}~\cite{zhu2021clusterability} and Li~\cite{li2021provably} introduced the alternatives to anchor points to learn the noise transition during the training. Jakramate \textit{et al.,}~\cite{bootkrajang2022towards} provides an elastic regularization method, brNet, for the general transition-based methods under small data, which integrates the prior belief on noise estimation via an adaptive sparse structure constraint. In Table~\ref{tab:transition_summary}, we summarize some popular transition-based methods and their learning strategies for noise transition.} Nevertheless, the update of noise transition as a noise adaptation layer through the back-propagation is ill-posed, since actually the noise transition should be up to the statistics of the complete dataset instead of an arbitrary min-batch of samples.

\begin{table}[t!]
    \centering
    \revision{
    \caption{Summary of recently popular transition-based methods.}
    \begin{tabular}{c|c}
       \hline
       \textbf{Method} & \textbf{Learning Strategy of transition}\\
       \hline
       Forward~\cite{patrini2017making} & Pre-estimation \\
       \hline
       S-Adaptation~\cite{goldberger2016training} & Pre-estimation \& Adaptation \\
       \hline
       T-revision~\cite{xia2019anchor} & Pre-estimation \& Residual Adaptation \\
       \hline
       Dual T~\cite{yao2020dual} & Divide-and-Conquer Pre-estimation \\
       \hline 
       HOC~\cite{zhu2021clusterability} & Clusterability \\
       \hline 
       VolMinNet~\cite{li2021provably} &  Anchor-free Learning \\
       \hline
       brNet~\cite{bootkrajang2022towards} & Regularized Noise Adaptation \\
       \hline
    \end{tabular}
    }
    \label{tab:transition_summary}
\end{table}

\subsection{Learning via Sample Re-weighting}
This line of works regulate the importance of each sample in the parameter estimation to reduce the noise effect~\cite{liu2016classification,pmlr-v80-ren18a,jiang2018mentornet,jiang2020beyond,wang2021learning}. It can be implemented in the perspectives of the label re-weighting or the sample pair re-weighting based on the modeling. For example, Reed \textit{et al.,}~\cite{reed2014training} leveraged the perceptual consistency to construct a surrogate of supervision by combining the label and the prediction, which showed the robustness to label noise. Li \textit{et al.,}~\cite{li2017learning} further substituted the prediction with the refined label by the graph distillation. Wang \textit{et al.,}~\cite{wang2018iterative} utilized the local intrinsic dimensionality to dynamically adapt the weight between the label and the prediction for Bootstrapping~\cite{reed2014training}. Jiang \textit{et al.,}~\cite{jiang2018mentornet,jiang2020beyond} designed a curriculum to measure the quality of each sample, and added the learned weights into the loss function to reduce the label noise effect. Shu \textit{et al.,}~\cite{shu2019metaweight} followed the similar motivation but constructed an explicit mapping to reweight the training samples under label noise. Recently, several works~\cite{NIPS2018_7454,Han2018CoteachingRT,yu2019does,wu2019collaborative,wu2021cooperative} explored to build two or more networks to collaboratively learn the weights for samples, showing the promise to the improvement of training. 

\subsection{Learning via Model Regularization}
This type of methods regularize the training process or the feature space to handle the adverse impact of label noise. Zhang et al.,~\cite{zhang2016understanding} discover DNNs can easily memorize the random labels completely, characterizing the challenge of deep learning with noisy labels. 
Their proposed Mixup~\cite{zhang2017mixup} that adopts the convex combinations of input features and noisy labels as the augmentation, has been demonstrated as an efficient regularization to prevent DNNs from noise-fitting. Arpit et al.,~\cite{arpit2017closer} investigated the memorization order of DNNs on noisy datasets and found the dropout could efficiently limit the speed of memorization on noise in DNNs. Tanaka et al.,~\cite{tanaka2018joint} explicitly introduced a regularization term to prevent the trivial case of assigning all labels to a single class in label correction. DivideMix~\cite{li2019dividemix} improves MixMatch~\cite{berthelot2019mixmatch} with label co-refinement and co-guessing to account for label noise after the dataset division. In the following sections, we will also show LCCN is easily extended to this counterpart and achieve the state-of-the-art performance on a range of datasets with different settings.

\revision{\subsection{Learning via Robust Loss Functions}
Loss functions are essential to train deep neural networks and the 0-1 loss has been demonstrated robust to label noise~\cite{hu2018does}. To integrated the robustness characteristic of the 0-1 loss and construct a differentiable counterpart, there are a range of works that explore the corresponding surrogate. GCE~\cite{zhang2018generalized} combines the mean absolute error (MAE) loss with cross-entropy loss to make a trade-off between robustness and optimization. Compared with the conventional surrogate losses, Lyv et al.,~\cite{lyu2019curriculum} introduced a tighter upper bound of the 0-1 loss, namely curriculum loss, which adaptively selects samples for stage-wise training. Ma et al.,~\cite{ma2020normalized} pointed out the dilemma of the loss design in pursuing robustness is that we may sacrifice the sufficient training and thus lead to the performance degeneration. To solve this problem, they showed a simple and efficient normalization can make the conventional losses more robust and introduced an active passive loss to achieve the robustness and the training efficiency. Liu et al.,~\cite{liu2020peer} proposed a new family of loss functions, peer loss functions, which originates from the peer prediction literature and the perspective of the Bayes optimal classifier. Nevertheless, the existing symmetric robust loss functions is restricted by the
symmetric condition, which is too stringent in optimization~\cite{zhou2021asymmetric}. Therefore, they developed asymmetric robust loss functions to relax this fitting difficulty. Overall, although robust loss functions are noise-tolerant to train DNNs, it is limited to only reduce the adverse effect of label noise in the head of the model.
}

\revision{\subsection{EM Optimization}
The EM algorithm has been a popular choice to optimize the statistical models, especially when the computational cost in the parameter space is affordable during the training. There are some classical works related to label noise, which follow the spirit of the EM optimization to learn the model parameters. For example, \cite{lawrence2001estimating} and \cite{bootkrajang2011multi} model the label noise respectively in the KFD for binary classification and rNDA for multi-class counterpart, and resort to the EM optimization to alternatively learn the instance-centric estimation of category probabilities and the class-centric parameters. Such an alternative optimization is reasonable and affordable to their generative probabilistic models with a few parameters (\textit{e.g.,} Gaussian mean and variance), but is hardly applicable to deep neural networks with millions of parameters. In the recent years, there are some neural network based works developed to combat the label noise with the EM optimization. For example, \cite{vahdat2017toward} targets to a more complex multi-label classification problem under the label noise, which represents the relationship between clean labels and noisy labels via conditional random fields. They leverage the Gibbs sampling plugged into EM to accelerate the optimization but also requires a clean auxiliary subset to calibrate the learning process. In \cite{wang2021robust}, although we are dealing with a different problem under the label noise, they share a similar idea with our work to learn the distribution of the latent variable via Collapse Gibbs sampling. Nevertheless, a noticeable difference in the scenario of \cite{wang2021robust} is their EM alternation (see the Algorithm 2 of \cite{wang2021robust}) might be a bottleneck if a very deep auto-encoder is used to model the document-topic distribution in their generative modeling. In this work, our proposed optimization method is not an standard EM algorithm but sounds like its stochastic version. Especially, our method decouples the model into two independent parts, the classifier component and the noise component, via a Gibbs sampler, and thus are not limited by the EM alternation and scalable in parallel.
}

\section{The Proposed Framework}
\subsection{Preliminaries}
In the $K$-class classification setting, a collection of $N$ training pairs $\{(x_n,\tilde{y}_n)\}_{n=1}^N$ is given, where $x_n$ \revision{are} the raw input data, e.g., the feature vector, and $\tilde{y}_n\in\{1,\dots,K\}$ is the corresponding noisy label. Assume $y_n$ denotes the true label of $x_n$ that is usually unknown in practice. Then, the goal in this task is to train a deep network classifier from the noisy dataset $\{(x_n,\tilde{y}_n)\}_{n=1}^N$ analogous to the training on the clean dataset $\{(x_n, y_n)\}_{n=1}^N$, so that it generalizes well on a clean test dataset regarding classification. As shown in~\cite{zhang2016understanding}, directly minimizing the following equation will make DNNs memorize both the clean samples and the noise,
\begin{align} \label{eq:pse_goal}
\hat{f}_\theta = \mathop{\text{argmin}}_{f_\theta\in\mathcal{F}}\frac{1}{N}\sum_{n=1}^N\ell(\tilde{y}_n,f_\theta(x_n)),
\end{align}
where $f_\theta$ is from the functional class $\mathcal{F}$ modeled by DNNs with the parameter $\theta$, and $\ell$ is the loss function to measure the discrepancy between the noisy label $\tilde{y}_n$ and the prediction $f_\theta(x_n)$. According to the universal approximation theorem~\cite{cybenko1989approximation,hornik1989multilayer}, neural networks have the capacity to model the arbitrarily complex data, even the random noise~\cite{zhang2016understanding}. That means, without any intervention in Eq.~\eqref{eq:pse_goal}, we will acquire a classifier that learns the random noise patterns.
To handle this issue, the transition-based approaches are proposed as the forward correction mechanism~\cite{liu2016classification,patrini2017making}, which constructs the following empirical risk minimization,
\begin{align} \label{eq:goal}
\hat{f}_\theta,\phi = \mathop{\text{argmin}}_{f_\theta\in\mathcal{F},\phi\in\Delta}\frac{1}{N}\sum_{n=1}^N\ell(\tilde{y}_n,\phi\circ f_\theta(x_n)).
\end{align}
Here $\phi$ is the noise transition matrix defined on the simplex $\Delta\in\mathbf{R}^{K\times K}$ and $f_\theta$ follows the same definition in Eq.~\eqref{eq:pse_goal}. 
The operator $\circ$ between them represents the \revision{function composition~\cite{shapiro2012composition}},
which indicates the mapping from the input to the noisy label is the sequential effect of both the prediction of the classifier and the noise transition. Compared with Eq.~\eqref{eq:pse_goal}, there is an extra term $\phi$ in Eq.~\eqref{eq:goal} to model the noise effect. Based on the Theorem 2 of~\cite{patrini2017making}, such a noise transition makes the training with the class-dependent noise achieve the same minimizer as the training with the clean data under the theory of the composite loss~\cite{reid2010composite}. More details can be referred to~\cite{reid2010composite,patrini2017making}. Although $\phi$ in Eq.~\eqref{eq:goal} could force the DNNs not to totally trust noisy labels, one critical problem is that it is hard to acquire an accurate estimate of $\phi$ in advance. This drives works~\cite{sukhbaatar2014training,goldberger2016training,xia2019t_revision,shu2019metaweight} that explore to progressively learn $\phi$ as a neural layer along with training. 

\subsubsection{The issue of stochastic training as a neural layer}
Let $\phi_{ij}=P(\tilde{y}=j|y=i)$ denote the transition from the category $y=i$ to the category $\tilde{y}=j$. When it is modeled by a neural layer with the parameter $W_{ij}$, we can formulate the popular optimization via the backpropagation under the mini-batch size of 1 as follows,
\begin{align}\label{eq:stissue}
\begin{split}
    & \revision{W_{ij}:=W_{ij} + \eta \frac{\partial\ell(\tilde{y}, z)}{\partial z}\frac{\partial z}{\partial W_{ij}}\Big|_{z=P(\tilde{y}|x)},}\\
\end{split}
\end{align}
where $\eta$ is the learning rate and \revision{$z=P(\tilde{y}|x)$} is the prediction. Given the typical loss $\ell$ in Eq.\eqref{eq:stissue}, \textit{e.g.,} the cross-entropy loss, \revision{$\frac{\partial\ell(\tilde{y},z)}{\partial  z}$} is usually unbounded if the prediction is arbitrarily divergent from the label. Thus, the gradient magnitude of $W_{ij}$ could be indefinitely large and in this case, $W_{ij}$ would be arbitrarily updated according to Eq.~\eqref{eq:stissue}. This \revision{induces training instability} especially in the worst case when we meet some extremely noisy samples near the optimal $W^*_{ij}$, which explains why we need to carefully adapt the transition layer in the training~\cite{goldberger2016training}. \revision{Note that, we can actually alleviate this if we have a larger mini-batch size. Specially, with the increasing mini-batch size, the estimated gradient in Eq.~\eqref{eq:stissue} becomes more accurate and stable to noise. However, such an effectiveness concomitantly induces an expensive computational cost, whose extreme is an computationally expensive EM~\cite{pmlr-v37-menon15} given in the following. This is usually undesired in the training of DNNs, when the computation resources are constrained. Therefore, our goal in this work is to construct an effective and efficient solution to address this dilemma both in the modeling and the optimization.}

\begin{figure} 
\centering
\includegraphics[width=0.48\textwidth]{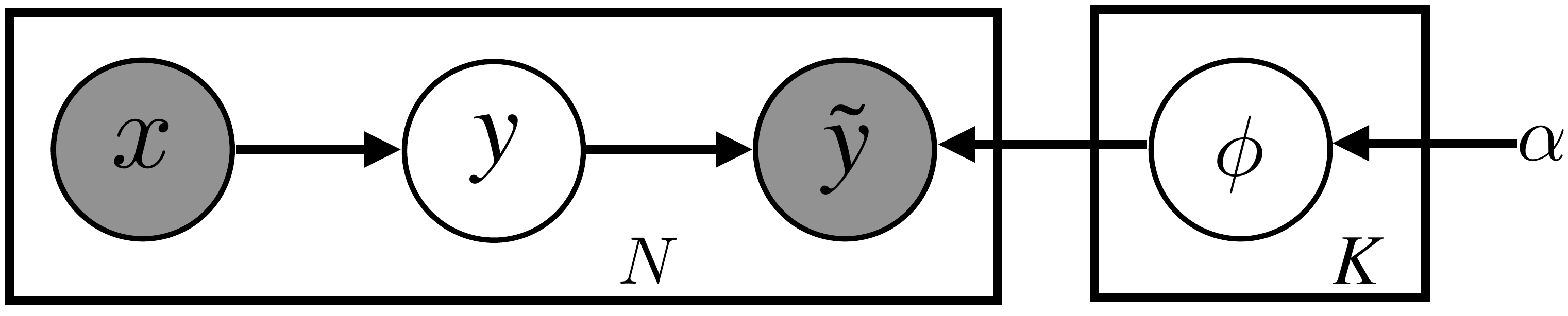}
\caption{Latent Class-Conditional Noise model. ($x$, $\tilde{y}$) is the sample pair. $y$ is the latent true label. $\phi$ is the noise transition. $\alpha$ is the Dirichlet hyperparameter. $N$ is the sample number and $K$ is the category number.}\label{fig:bccn}
\end{figure}

\subsection{Latent Class-Conditional Noise model}
In this section, we will present a Latent Class-Conditional Noise (LCCN) model, 
which characterizes noise transition under a Bayesian framework. The intuition behind LCCN is explicitly endowing the noise transition with the statistics of the complete dataset, which constrains the optimization unlike the backpropgation in Eq.~\eqref{eq:stissue}, avoiding the arbitrary update from the mini-batch of samples.
Fig.~\ref{fig:bccn} illustrates its graphical model and the generative process is as follows,
\begin{itemize}
\item The latent label $y_n\sim P(\cdot|x_n)$, where $P(\cdot|x_n)$ is a \emph{Categorical} distribution modeled by the deep neural network $f_\theta$ and the given $x_n$ is its input feature.
\item The transition vector of the $k$th class $\phi_k\sim P(\cdot|\alpha)$, where $\alpha$ is the parameter of a $Dirichlet$ distribution $P(\cdot|\alpha)$ and $\left[\phi_1,\cdots,\phi_K\right]^T$ is the noise transition.
\item The noisy label $\tilde{y}_n\sim P(\cdot|\phi_{y_n})$, where $P(\cdot|\phi_{y_n})$ is a \emph{Categorical} distribution parameterized by $\phi_{y_n}$.
\end{itemize}

The above generative process explicitly constrains the noise transition variable $\phi$ responsible for the generation of all samples $\{(x_n, \tilde{y}_n)\}_{n=1,\dots,N}$. This actually builds the dependency relationship between the estimation of noise transition and the statistics of all samples instead of a mini-batch of samples. Specifically, different from the modeling by the neural layer that is optimized by the derivatives of the mini-batch of samples, we must optimize $\phi$ onto the Dirichlet-distributed space characterized by the complete dataset. Compared to previous works, such a difference makes us avoid the unstable issue in Eq.~\eqref{eq:stissue} when jointly trained with DNNs. However, we have to admit this difference also introduces the difficulty to efficiently optimize the LCCN model with the current off-the-shelf methods in the context of deep learning. \revision{Specially, the general idea is to apply amortized variational inference~\cite{kingma2013auto,MnihNVI}, which induces expensive EM updates. Let $(X, Y, \widetilde{Y})$ denote the collection of all true and noisy variables of $N$ sample triplets We give the following deduction to characterize the problem.
\begin{align} \label{eq:EM_obj}
\begin{split}
    & \ln{P(\widetilde{Y}|X;\alpha)}  \\
    & = \ln{\int P(\phi;\alpha)\prod_{n=1}^N\left(\sum_{y_n=1}^KP(\tilde{y}_n|y_n,\phi)P(y_n|x_n)\right)} \mathrm{d}\phi \\
    & \geq \int P(\phi;\alpha) \sum_{n=1}^N\sum_{y_n=1}^K \phi_{y_n\tilde{y}_n}\ln{P(y_n|x_n)}\mathrm{d}\phi \\
    & = \sum_{n=1}^N\sum_{y_n=1}^K \left(\int P(\phi;\alpha)  \phi_{y_n\tilde{y}_n}\mathrm{d}\phi\right) \ln{P(y_n|x_n)} \\
    & = \sum_{n=1}^N\sum_{y_n=1}^K \underbrace{\frac{\alpha_{y_n\tilde{y}_n}}{\sum_{k=1}^K \alpha_{y_nk}}}_{\bar{\phi}_{y_n\tilde{y}_n}} \ln{P(y_n|x_n)}. \\ 
\end{split}
\end{align}
Here, we use a new notation $\bar{\phi}$ (can be considered as the expected noise transition) in last line of Eq.~\eqref{eq:EM_obj} to completely represent the $\alpha$ proportion. Then, maximizing the above lower bound leads to the following EM updates.
\begin{align}
\begin{cases}
\textbf{E-Step:}~ \bar{\phi}_{kk'}= \frac{\sum_{n=1}^N \mathbbm{1}\left(\tilde{y}_n=k'\right)P(y_n=k|x_n)}{\sum_{n=1}^N P(y_n=k|x_n)}\\
\textbf{M-Step:}~ \max~ \sum_{n=1}^N\sum_{y_n=1} \bar{\phi}_{y_n\tilde{y}_n}\ln{P(y_n|x_n)} \\
\end{cases}. \nonumber 
\end{align} 
Such an alternative optimization is usually computationally expensive, since it requires the  computation of all samples in the dataset to estimate  $\bar{\phi}$~\cite{goldberger2016training} in each epoch. Considering this, we deduce a new more efficient dynamic label regression method for LCCN inspired by Gibbs sampling~\cite{griffiths2004finding}.}

\subsection{Dynamic Label Regression}
The proposed dynamic label regression for LCCN consists of two parts, \emph{Autoencoded Gibbs sampling} and \emph{loss minimization}. It alternates between the \emph{Autoencoded Gibbs sampling} to infer the latent true labels and the \emph{Loss Minimization} for parameter learning. In the following, we show its deduction procedure by a two-step formulation.  Note that, \revision{different from the above EM updates~\cite{dempster1977maximum}, our optimization resembles the simple forward-backward pass in deep learning}, and compared to previous noise-transition-based methods, it only introduces a little of extra computational cost, which will be analyzed in the following sections.

\subsubsection{Autoencoded Gibbs sampling}
This method is \revision{roughly in line with Gibbs sampling applied to the Dirichlet-Multinomial conjugation~\cite{griffiths2004finding}}. According to the methodology, we need to first compute the posterior of $y$ based on its variable-dependent path of the graphical model in Fig.~\ref{fig:bccn} as follows,
\begin{align} \label{eq:step1}
\begin{split}
& P(Y|X,\widetilde{Y};\alpha) \\ 
& = \int_\phi \prod_{k=1}^K P(\phi_k;\alpha) \prod_{n=1}^N P(y_n|x_n,\tilde{y}_n,\phi) d\phi \\
& = \int_\phi \prod_{k=1}^K P(\phi_k;\alpha) \prod_{n=1}^N \frac{P(y_n|x_n)P(\tilde{y}_n|y_n,\phi)}{P(\tilde{y}_n|x_n)} d\phi \\
& = \underbrace{\prod_{n=1}^N \frac{P(y_n|x_n)}{P(\tilde{y}_n|x_n)}}_{\text{integral-free part}} \int_\phi \underbrace{\prod_{k=1}^K P(\phi_k;\alpha) \prod_{n=1}^N P(\tilde{y}_n|y_n,\phi)}_{\text{integral-related part}} d\phi \\
& = S  \int_\phi \prod_{k=1}^K \frac{\Gamma(\sum_{k'}^K \alpha_{k'})}{\prod_{k'}^K \Gamma(\alpha_{k'})} \prod_{k'}^K \phi_{kk'}^{\alpha_{k'}-1}  \prod_{n=1}^N \phi_{y_n \tilde{y}_n} d\phi,
\end{split}
\end{align}
where $S=\prod_{n=1}^N \frac{P(y_n|x_n)}{P(\tilde{y}_n|x_n)}$ \revision{is} to simplify the equation and will be canceled out finally. The last line of Eq.~\eqref{eq:step1} is acquired by substituting the probability with the corresponding distribution. 
\revision{Note that, the goal in deducing $P(Y|X,\widetilde{Y};\alpha)$ is not to minimize one objective but to estimate the posterior distribution of the true label $y$.}

Regarding Eq.~\eqref{eq:step1}, if we use $\revision{C}_{(\cdot)(\cdot)}$ to represent the confusion matrix of the noisy dataset, we can have $\sum_k^K\sum_{k'}^K \revision{C}_{kk'}$=$N$ and $\prod_{n=1}^N \phi_{y_n \tilde{y}_n} = \prod_k^K\prod_{k'}^K \phi_{kk'}^{\revision{C}_{kk'}}$. Here, we introduce this new notation to simplify the explanations in the verbose equations. Putting it into Eq.~\eqref{eq:step1} and applying the conjugation between the Dirichlet distribution and the Multinominal distribution, we rewrite Eq.~\eqref{eq:step1} as follows,
\begin{align} \label{eq:step2}
& P(Y|X,\widetilde{Y};\alpha)  \nonumber \\
& = S  \int_\phi \prod_{k=1}^K \frac{\Gamma(\sum_{k'}^K \alpha_{k'})}{\prod_{k'}^K \Gamma(\alpha_{k'})} \prod_{k'}^K \phi_{kk'}^{\revision{C}_{kk'}+\alpha_{k'}-1} d\phi \nonumber\\
& = S  \prod_{k=1}^K \frac{\Gamma(\sum_{k'}^K \alpha_{k'})}{\prod_{k'}^K \Gamma(\alpha_{k'})} \prod_{k=1}^K \frac{\prod_{k'}^K \Gamma(\alpha_{k'}+\revision{C}_{kk'})}{\Gamma(\sum_{k'}^K(\alpha_{k'}+\revision{C}_{kk'}))}.  \\
& = \prod_{n=1}^N \frac{P(y_n|x_n)}{P(\tilde{y}_n|x_n)} \prod_{k=1}^K \frac{\Gamma(\sum_{k'}^K \alpha_{k'})}{\prod_{k'}^K \Gamma(\alpha_{k'})} \prod_{k=1}^K \frac{\prod_{k'}^K \Gamma(\alpha_{k'}+\revision{C}_{kk'})}{\Gamma(\sum_{k'}^K(\alpha_{k'}+\revision{C}_{kk'}))}.\nonumber
\end{align}
\revision{However, the probability computation of $P(Y|X,\widetilde{Y};\alpha)$ is about all instances and cannot be decoupled due to the cumulative product and the coupled $\revision{C}_{(\cdot)(\cdot)}$. That is to say, according to Eq.~\eqref{eq:step2}, every time, we have the $\mathcal{O}(NK)$ complexity to compute a sampling probability $P(Y|X,\widetilde{Y};\alpha)$ from a very high-dimensional simplex space $\mathbf{R}^{N(K-1)}$. Fortunately, there is an efficient solution to this issue via Gibbs sampling like the sampler for LDA~\cite{griffiths2004finding}}. According to the Gibbs sampling, we only need to compute $P(y_n|Y^{\neg n})$ first\footnote{Note that $\neg$ means removing the current object statistic from the whole collection of all object statistics.}, and then based on $P(y_n|Y^{\neg n})$, sample a sequence of observations. The following deduction leverages Eq.~\eqref{eq:step2} and $\Gamma(x+1)=x\Gamma(x)$ to reach our target sampling probability.
\begin{align} \label{eq:step3}
& P(y_n|Y^{\neg n},X,\widetilde{Y};\alpha)  \nonumber \\
& = \frac{P(\revision{Y=\{y_n\}\cup Y^{\neg n}}|X,\widetilde{Y};\alpha)}{P(Y^{\neg n}|X,\widetilde{Y};\alpha)} \nonumber \\
& = \revision{\frac{\prod_{n'=1}^N \frac{P(y_{n'}|x_{n'})}{P(\tilde{y}_{n'}|x_{n'})} \prod_{k=1}^K \frac{\Gamma(\sum_{k'}^K \alpha_{k'})}{\prod_{k'}^K \Gamma(\alpha_{k'})} \prod_{k=1}^K \frac{\prod_{k'}^K \Gamma(\alpha_{k'}+C_{kk'})}{\Gamma(\sum_{k'}^K(\alpha_{k'}+C_{kk'}))}}{\prod_{n'\neq n} \frac{P(y_{n'}|x_{n'})}{P(\tilde{y}_{n'}|x_{n'})} \prod_{k=1}^K \frac{\Gamma(\sum_{k'}^K \alpha_{k'})}{\prod_{k'}^K \Gamma(\alpha_{k'})} \prod_{k=1}^K \frac{\prod_{k'}^K \Gamma(\alpha_{k'}+C_{kk'}^{\neg n})}{\Gamma(\sum_{k'}^K(\alpha_{k'}+C_{kk'}^{\neg n}))}}} \nonumber \\
& = \frac{P(y_n|x_n)}{P(\tilde{y}_n|x_n)} \frac{\alpha_{\tilde{y}_n}+\revision{C}_{y_n\tilde{y}_n}^{\neg n}}{\sum_{k'}^K(\alpha_{k'}+\revision{C}_{y_n k'}^{\neg n})}  \\
&\propto \underbrace{P(y_n|x_n)}_{\text{Classifier encoder}} \underbrace{\frac{\alpha_{\tilde{y}_n}+\revision{C}_{y_n \tilde{y}_n}^{\neg n}}{\sum_{k'}^K(\alpha_{k'}+\revision{C}_{y_n k'}^{\neg n})}}_{\text{Conditional transition}}. \nonumber
\end{align}
For the last line of Eq.~\eqref{eq:step3}, the proportionally scaling is used to remove the constant $P(\tilde{y}_n|x_n)$ for any $y_n$, since it does not affect the sampling but helps reduce the computation.
\revision{In general, $P(y_n|Y^{\neg n},X,\widetilde{Y};\alpha)$ is the posterior label estimation for $x_n$ on the basis of the prior label estimation $P(y_n|x_n)$ and the \emph{conditional transition} term in the last line of Eq.~\eqref{eq:step3}). $P(y_n|x_n)$ is the current model probabilistic estimate, which is trained based on $x_n$ and its associated $y_n$ sampled from the early Gibbs chains, instead of $x_n$ and $\tilde{y}_n$. It is continually evolving to approach the ideal model trained with the clean dataset, since the Gibbs sampling progressively allows us to sample the accurate $y_n$ of $x_n$ for training. Simultaneously, the conditional transition term that encodes the knowledge of noise transition, acts as a re-weighting regularization to make the noisy observation well explained. Note that, the conditional transition is also progressively refined based on the category allocation in $N^\neg_{(\cdot)(\cdot)}$ updated with the Gibbs sampling, and can be roughly considered as the role of the posterior distribution over $\phi$. Given the above analysis, we specially term the deduced method as \emph{Autoencoded Gibbs sampling} and summarize its procedure in Algorithm~\ref{alg:ags}\footnote{We present the stochastic-training version in the mini-batch level.}.}

\subsubsection{Loss Minimization} According to  Eq.~\eqref{eq:step3}, we can sample a collection of posterior latent true labels $\{z\}$. Such new observations then naturally decompose the optimization of LCCN into two parts, modeling the noise modeling and training the classifier, which are formulated as the following two sub-problems,
\begin{align} \label{eq:minimization}
\begin{cases}
\min \revision{-\frac{1}{M}\sum_{m=1}^M \ell_1(y_m, P(y_m|x_m))} \\ 
\min -\frac{1}{N}\sum_{n=1}^N \ell_2(\tilde{y}_n,P(\tilde{y}_n|y_n)), 
\end{cases}
\end{align}
where $M$ is the mini-batch size, $N$ is the dataset size, $\ell_1$ is the $\xi$-clipped cross-entropy loss\footnote{The prediction from the model is clipped between $\xi$ and $1-\xi$ for the sampling probabilities on all possible categories, e.g., $\xi=10^{-20}$.} and $\ell_2$ is the negative log-likelihood loss.
The upper objective in Eq.~\eqref{eq:minimization} utilizes the posterior latent true label $y$ instead of $\tilde{y}$ to supervise the training of the classifier. The lower objective optimizes the noise modeling by learning the transition. Note that, for the lower objective, we apparently leverage the whole set of $y$ for each sample, but actually only locally replace the early assigned $\{y_n^{\text{old}}\}$ of the mini-batch of samples with the newly sampled latent true labels $\{y_n^{\text{new}}\}$ in the current iteration. This induces the partially statistic update of $\revision{C}_{(\cdot)(\cdot)}$ regarding the current mini-batch of samples, with which, the negative log-likelihood loss has the following solution, 
\begin{align}
    \begin{split}
        & \phi_{kk'} = \frac{\revision{C}_{kk'}}{\sum_{k'}\revision{C}_{kk'}}.
    \end{split}
\end{align}
In the global view, it indicates the noise transition depends on the dynamic allocation statistic $\revision{C}_{(\cdot)(\cdot)}$ about the whole dataset, and thus would not suffer from the rapid update as in the backpropagation in~\cite{goldberger2016training}. The optimization for the classifier and the transition is summarized in Algorithm~\ref{alg:lm}.

\begin{algorithm}[t]
  \begin{algorithmic}[1]
  \REQUIRE The DNN classifier $f_\theta$, the confusion matrix $\revision{C}_{(\cdot)(\cdot)}$ and a mini-batch of samples $\{(x_m, \tilde{y}_m)\}_{m=1}^M$.
  \STATE forward the samples to $f_\theta$ to compute $\{f_\theta(x_m)\}_{m=1}^M$.
  \FOR{$m=1,\dots, M$}
  \STATE sample $z_m$ based on $f_\theta(x_m)$ and $N^\neg_{(\cdot)(\cdot)}$ by Eq.~\eqref{eq:step3}.
  \ENDFOR
  \RETURN a mini-batch of labels $\{y_m\}_{m=1}^M$.
  \end{algorithmic}
  \caption{Autoencoded Gibbs sampling}
  \label{alg:ags}
\end{algorithm}
\begin{algorithm}[t]
  \begin{algorithmic}[1]
  \REQUIRE The DNN classifier $f_\theta$, the confusion matrix $\revision{C}_{(\cdot)(\cdot)}$ and a mini-batch of samples $\{(x_m, y_m, \tilde{y}_m)\}_{m=1}^M$.
  \STATE compute the loss $\ell_1$ based on $\{(f_\theta(x_m), y_m)\}_{m=1}^M$.
  \STATE update the classifier $\theta:=\theta-\lambda\frac{1}{M}\sum_{m=1}^{M}\nabla_{\theta}\ell_1$ .
  \STATE update the matrix $\revision{C}_{(\cdot)(\cdot)}$ based on $\{(y_m, \tilde{y}_m)\}_{m=1}^M$.
  \RETURN the classifier $f_\theta$ and the confusion matrix $\revision{C}_{(\cdot)(\cdot)}$.
  \end{algorithmic}
  \caption{Loss Minimization}
  \label{alg:lm}
\end{algorithm}

\subsubsection{The Complete Algorithm and Complexity} 
Iterating the procedure of Eq.~\eqref{eq:step3} and Eq.~\eqref{eq:minimization}, we gradually infer the latent true label, train the classifier and estimate the noise transition as shown in Fig.~\ref{fig:network}. Especially in this progressive procedure, when $P(y|x)$ approaches the true distribution of clean labels, the training of the classifier is similar to that on the clean dataset. We will give the analysis using Theorem 2 in the subsequent section. \revision{Note that, one practical drawback  in the implementation} for DNNs is the cold-start problem at the early stage of training. Following~\cite{goldberger2016training}, we include \revision{a following warming-up transition,}
\begin{align} \label{eq:cold-start}
\phi^{\text{init}}_{kk'} &= \frac{\Sigma_n^N \mathbbm{1}\left(\tilde{y}_n=k'\right)P(y_n=k|x_n)}{\Sigma_n^N P(y_n=k|x_n)}.
\end{align}
 The complete algorithm to stack \emph{Autoencoded Gibbs sampling}, \emph{Loss Minimization} as well as the warming-up for LCCN \revision{is} all summarized in Algorithm~\ref{alg:dlr}.
We term this algorithm as \emph{dynamic label regression} under noisy labels to mean that the latent true label is progressively learned in the interaction dynamics between Algorithm~\ref{alg:ags} and Algorithm~\ref{alg:lm}, and finally we train the classifier and learn the noise transition.

\revision{Regarding the complexity,} the backpropagation optimization of a DNN model involves two steps, the forward and the backward computations. In each mini-batch update, its time complexity is $\mathcal{O}(M\Lambda)$, where $M$ is the mini-batch size and $\Lambda$ is the parameter size. Here, in Algorithm~\ref{alg:dlr}, we additionally add a sampling operation via Eq.~\eqref{eq:step3} whose complexity is $\mathcal{O}(M+K^2)$ ($K$ is the category number). Note that, the first term in the RHS of Eq.~\eqref{eq:step3} has been computed in the forward pass. Since $M$ and $K$ is usually significantly smaller than the million-scale $\Lambda$, \textit{i.e.,} $\max(M, K) \ll \Lambda$, the computational cost for the sampling is negligible compared to $\mathcal{O}(M\Lambda)$. Besides, the optimization for noise modeling in Eq.~\eqref{eq:minimization} can be ignored, as it only involves the simple normalization of a confusion matrix without other cost for derivatives in the neural layers. In total, the big-O complexity of each mini-batch remains the same and our method is scalable to big data. 

\begin{algorithm}[t]
  \begin{algorithmic}[1]
  \REQUIRE A noisy dataset $\mathcal{D}=\{(x, \tilde{y})\}$, a classifier $P(\cdot|x)$ modeled by $f_\theta$, warming-up steps $\delta$, the number of iterations $L$ and the mini-batch size $M$.
  \STATE pretrain the classifier $f_\theta$ on the noisy dataset $\mathcal{D}$.
  \STATE compute the warming-up noise transition matrix $\phi^{\text{init}}$.
  \STATE compute the confusion matrix $\revision{C}_{(\cdot)(\cdot)}$.
  \FOR{$l=1,\dots,L$}
  \STATE \revision{sample} a mini-batch $\{(x_m, \tilde{y}_m)\}_{m=1}^M$.
  \IF{$l<\delta$}
  \STATE Replace the conditional transition in Eq.~\eqref{eq:step3} by $\phi^{\text{init}}$.
  \ENDIF
  \STATE $\{y_m\}_{m=1}^M$ = Algorithm~\ref{alg:ags}$\left(f_\theta, \revision{C}_{(\cdot)(\cdot)}, \{(x_m, \tilde{y}_m)\}_{m=1}^M\right)$.
  \STATE $f_\theta, \revision{C}_{(\cdot)(\cdot)}$ = Algorithm~\ref{alg:lm}$\left(f_\theta, \revision{C}_{(\cdot)(\cdot)}, \{(x_m, y_m, \tilde{y}_m)\}_{m=1}^M\right)$.
  \ENDFOR
  \STATE Output the classifier $f_\theta$ and the noise transition $\phi$.
  \end{algorithmic}
  \caption{Dynamic Label Regression for LCCN}
  \label{alg:dlr}
\end{algorithm}
  
\subsection{Theoretical analysis}
In this section, we provide the theoretical analysis of the dynamic label regression method for LCCN in the perspective of the mixing time to Gibbs sampling and the \revision{generalization} bound of learning with noisy labels.
\begin{lemma}
For a reversible, irreducible and aperiodic Markov chain with the state space $\Omega$, let \revision{$\textbf{Q}\in \mathbb{R}^{K\times K}$ be the state transition matrix describing the transition probability from one estimation of latent true label $Y$ to the other estimation of $Y$, $\pi$ is the underlying stationary distribution \revision{of $Y$}, \textit{i.e.,} $\pi=\pi\textbf{Q}$.} \revision{Let $\pi_{min}=\min_{Y\in\Omega}\pi(Y)$ and $\lambda^*$ be the maximal absolute eigenvalue of the state transition matrix \revision{\textbf{Q}}}. Then, the $\epsilon$-mixing time from the initial arbitrary state to the equilibrium is characterized by the following bounds,
\begin{align}\label{eq:lemma}
\frac{\lambda^*}{1-\lambda^*}\ln{\left(\frac{1}{2\epsilon}\right)}\leq \tau_{mix}(\epsilon)\leq \frac{1}{1-\lambda^*}\ln{\left(\frac{1}{\pi_{min}\epsilon}\right)},
\end{align}
where $\tau_{mix}(\epsilon)=\min\{t:||P_t(Y)-\pi||_{TV}\leq \epsilon\}$ and $||\cdot||_{TV}$ is the total variation distance between two probability measures.
\end{lemma} 
The above lemma indicates \revision{how many steps for Algorithm~\ref{alg:dlr} (more concretely for Eq.~\eqref{eq:step3} in Algorithm~\ref{alg:ags}) are required to estimate the stationary latent true labels for each sample, which}
is at most constantly linear to the inverse of $1-\lambda^*$. Note that, although it is hard to accurately quantify $\lambda^*$ due to the evolving state transition matrix~\cite{levin2017markov}, the recent work~\cite{johanmixing} discovers Gibbs sampling is efficient enough and almost \revision{has the complexity of $\mathcal{O}(\log{N})$}. 
\revision{In Fig.~\ref{fig:transition} and Fig.~\ref{fig:label_trace}, we will visualize how the noise transition is estimated and illustrate the label inference procedure in Algorithm~\ref{alg:dlr}.}

In statistical learning theory, the \emph{excess risk}~\footnote{\url{https://en.wikipedia.org/wiki/Risk_difference}} 
and \emph{the error bound} \textit{w.r.t.} the expected risk and Bayes risk, are two important quantities to measure model the generalization performance. In learning with noisy labels, such two quantities are bounded by the following generalization bound (see the Appendix A for the details) $$\Delta_\mathcal{F}=\sup_{f_\theta\in\mathcal{F}}\big|\mathbf{E}\left[\ell_1(y,f_\theta(x))\right]-\mathbf{E}^{(D_N)}\left[\ell_1(y, f_\theta(x))\right]\big|,$$ where $\mathbf{E}\left[\cdot\right]$ and $\mathbf{E}^{(D_N)}\left[\cdot\right]$ respectively represents the expectation on the clean data distribution and the empirical estimation with the data whose labels are from the Gibbs sampling. Therefore, analyzing the upper bound of $\Delta_\mathcal{F}$ can help understand which factors affect the generalization performance of LCCN. Formally, we deduce the following theorem to interpret some potential confounding factors.
\begin{theorem} \label{theorem}
Assume $f^*_\theta$ and $f^\dagger_\theta$ respectively are the underlying groundtruth labeling functions $\mathcal{X}\rightarrow\mathcal{Y}$ of clean test data and data from the Gibbs sampling. Define the composite function class $\mathcal{G}=\{x\mapsto \ell_1(f'_\theta(x),f_\theta(x)):f'_\theta,f_\theta\in\mathcal{F}\}$. Then, for any probability $\delta>0$, with probability at least $1-\delta$,
\begin{align}\label{eq:theorembound}
\begin{split}
& \Delta_\mathcal{F} \leq \Delta + \widehat{\mathcal{R}}(\mathcal{G}) + 3\rho \sqrt{\frac{\ln(\frac{2}{\delta})}{2N}},
\end{split}
\end{align}
where $\Delta=\sup_{f_\theta\in\mathcal{F}}\bigg|\mathbf{E}\left[\ell_1(f^*_\theta(x)-f^\dagger_\theta(x),f_\theta(x))\right]\bigg|$, $\widehat{\mathcal{R}}(\mathcal{G})$ is the Rademacher complexity~\cite{bartlett2002rademacher} of $\mathcal{G}$ and $\rho$ is the maximum of the $\xi$-clipped cross entropy loss, i.e., $-\ln{\xi}$.
\end{theorem}
The above theorem indicates the generalization performance of the classifier learned by LCCN depends upon three factors, i.e., the inherent gap $\Delta$ between the domain of the noisy training data and the domain of the clean test data, the function complexity $\widehat{\mathcal{R}}(\mathcal{G})$ and the sample number $N$. \revision{Note that, generally speaking, for \textit{class-conditional noise}, \textit{instance-dependent noise} and \textit{out-of-distribution noise}, the common factor is $\Delta$ in Eq.~\eqref{eq:theorembound}. For the former two types of noise, if LCCN can infer all the latent true labels of noisy data, we will have $f^*_\theta = f^\dagger_\theta$ and $\Delta=0$. Then, Eq.~\eqref{eq:theorembound} will degenerate to the Rademacher bound~\cite{mansour2009domain} after scaling the loss to $[0,1]$, and same to the training on the clean data. However, for out-of-distribution noise, it is impossible to completely remove the domain bias, since there are some indefinite samples different from that of the clean test data and non-identifiable by the pre-defined labels.} For example, the web data may contain many out-of-distribution categories. Thus, $\Delta$ is the bottleneck factor to the generalization performance in the real-world scenarios.

In the previous section, we analyzed the issue of stochastic training of \revision{noise transition} as the neural layer~\cite{goldberger2016training}. Here, we show that our method safeguards the update of the noise transition during the training via a mini-batch of samples. 
\begin{theorem}\label{theorem:update}
Suppose $\alpha_i$ is a positive smoothing scalar, $\revision{O}_i$ is the current sample number of the $i$th category (i=1,$\dots$,$K$), $\revision{T}_i$ is the sum of the sample numbers newly allocated into (positive) and removed from (negative) the $i$th category after a mini-batch of training samples, and $\revision{\widehat{T}}_i$ is its absolute sum of such two cases. Then, for the transition vector $\phi_i$ of the $i$th category, its variation via a training batch is characterized by the following equation,
\begin{align} \label{eq:theorem}
\big|\phi_i^{new}-\phi_i^{old}\big| \leq \frac{|r_i| + \widehat{r}_i}{1+r_i}
\end{align}
where $r_i=\frac{\revision{T}_i}{\revision{O}_i+\sum_{j=1}^K \alpha_j}$ and $\widehat{r}_i=\frac{\revision{\widehat{T}}_i}{\revision{O}_i+\sum_{j=1}^K \alpha_j}$. According to the definition, we have $r_i>-1$, $\widehat{r}_i\geq 0$ and $\widehat{r}_i\geq |r_i|$.
\end{theorem}
\begin{proof}
The variation of $\phi_i$ after a training batch is,
\begin{align}
\begin{split}
& \big|\phi_i^{new}-\phi_i^{old}\big|\\
& =\sum_{j=1}^K\big|\phi_{ij}^{new}-\phi_{ij}^{old}\big| \\
& = \sum_{j=1}^K\Bigg|\frac{\revision{O}_{ij}+\alpha_j+\revision{T}_{ij}}{\revision{O}_i+\sum_{j'=1}^K \alpha_{j'} + \revision{T}_i}-\frac{\revision{O}_{ij}+\alpha_j}{\revision{O}_i+\sum_{j'=1}^K \alpha_{j'}}\Bigg| \\
& \leq \sum_{j=1}^K\frac{\big|(\revision{O}_i+\sum_{j'=1}^K \alpha_{j'})\revision{T}_{ij}\big|+\big|(\revision{O}_{ij}+\alpha_j)M_i\big|}{(\revision{O}_i+\sum_{j'=1}^K \alpha_{j'})(\revision{O}_i+\sum_{j'=1}^K \alpha_{j'} + \revision{T}_i)} \\
& = \frac{(\revision{O}_i+\sum_{j'=1}^K \alpha_{j'})\revision{\widehat{T}}_i+(\revision{O}_i+\sum_{j'=1}^K \alpha_{j'})\big|\revision{T}_i\big|}{(\revision{O}_i+\sum_{j'=1}^K \alpha_{j'})(\revision{O}_i+\sum_{j'=1}^K \alpha_{j'} + \revision{T}_i)}\\
& = \frac{|r_i| + \widehat{r}_i}{1+r_i}
\end{split}
\end{align}
\end{proof}
\begin{corollary}~\label{corollary}
$M$ is the batch size in the training. If it satisfies the condition $M \ll \revision{O}_i$, we have $\widehat{r}_i < \frac{\revision{\widehat{T}}_i}{\revision{O}_i} < \frac{M}{\revision{O}_i}$ in a small scale. Then the variation of $\phi_i$ after a training mini-batch will be bounded by $\frac{|r_i| + \widehat{r}_i}{1+r_i}\leq \frac{2 \widehat{r}_i}{1-\widehat{r}_i}\approx 2 \widehat{r}_i$ in a small scale.
\end{corollary}
Theorem~\ref{theorem:update} and Corollary~\ref{corollary} demonstrate the update of the noise transition is bounded in each training iteration. Especially, the change becomes smaller with the decrease of the mini-batch size $M$, \textit{i.e.,} more stable. As aforementioned discussion, the core drawback in~\cite{goldberger2016training} is that the transition layer can be arbitrarily updated via a mini-batch of samples. Then, the noise transition might be pushed into a undesired local minimum by some extremely noisy training samples. The experimental parts will confirm this point. Conversely, the theory indicates this problem in LCCN is avoided by our optimization method. With this advantage, the conditional transition in Eq.~\eqref{eq:step3} could progressively change towards at the groundtruth when the classifier is well trained. Similarly, with more reliable labels in Eq.~\eqref{eq:minimization}, the classifier can be better learned and the noise modeling will be refined. As a result, we acquire a virtuous cycle for optimization.  

\subsection{Extensions to More Generalized Settings.}
In this section, we extend LCCN in Fig.~\ref{fig:bccn} to some generalized variants in Fig.~\ref{fig:outlier_semi},
which shares the optimization but might be more practical in real-world applications.

\subsubsection{Extension for Open-Set Noisy Labels} 
The dataset collected from online websites or real-world scenarios, might contain the open-set label noise~\cite{wang2018iterative}, \textit{i.e.,} have some abnormal categories of samples out of the pre-defined classes. However, previous class-conditional noise models~\cite{reed2014training,xiao2015learning,sukhbaatar2014training,patrini2017making,goldberger2016training} including our LCCN, mainly focus on the closed-set label corruption~\cite{wang2018iterative}, which cannot handle such OOD data. For example, it is hard to estimate the transition matrix via the mentioned two-step formulation in~\cite{patrini2017making}, since we have to select from the indefinite and infinite abnormal samples to construct the critical anchor set. Thus, it will be more useful if we can extend LCCN to handle the open-set noisy labels. Actually, one straightforward modification to LCCN is adding one extra dimension for the latent variable $y$, \textit{i.e.,} $y\in\{1,...,K, K+1\}$, to cater for the abnormal samples. Correspondingly, the noise transition is changed from $\mathbf{R}^{K\times K}$ to $\mathbf{R}^{(K+1)\times K}$. The $(K+1)$th index collapses OOD samples as outlier and seamlessly integrates into LCCN. We terms this construction as LCCN$^*$ and presents its graphical model in Fig.~\ref{fig:outlier_semi}(a). 
Note that, LCCN$^*$ does not alter the previous deduction and optimization. \revision{Besides, we would like to clarify that our aim here is not to compare with some open-set recognition methods~\cite{geng2020recent}, since they require much modification in the methodology, but to show a potential gain compared with previous methods that ignore the open-set label noise when it exists. The gain can be raised if we have more prior knowledge available, \textit{e.g.,} a clean subset, which we validate in the experiments. Besides, the design of LCCN$^*$ is orthogonal to OOD detection~\cite{yang2021oodsurvey} and open-set recognition~\cite{geng2020recent}, and the comprehensive extensions following such two paradigms can be more useful to cope with open-set noise, which we leave in future works.}

\subsubsection{Extension for Semi-supervised Learning} 
\begin{figure*} 
\centering
\subfigure[LCCN$^*$]{
\includegraphics[width=0.3\textwidth]{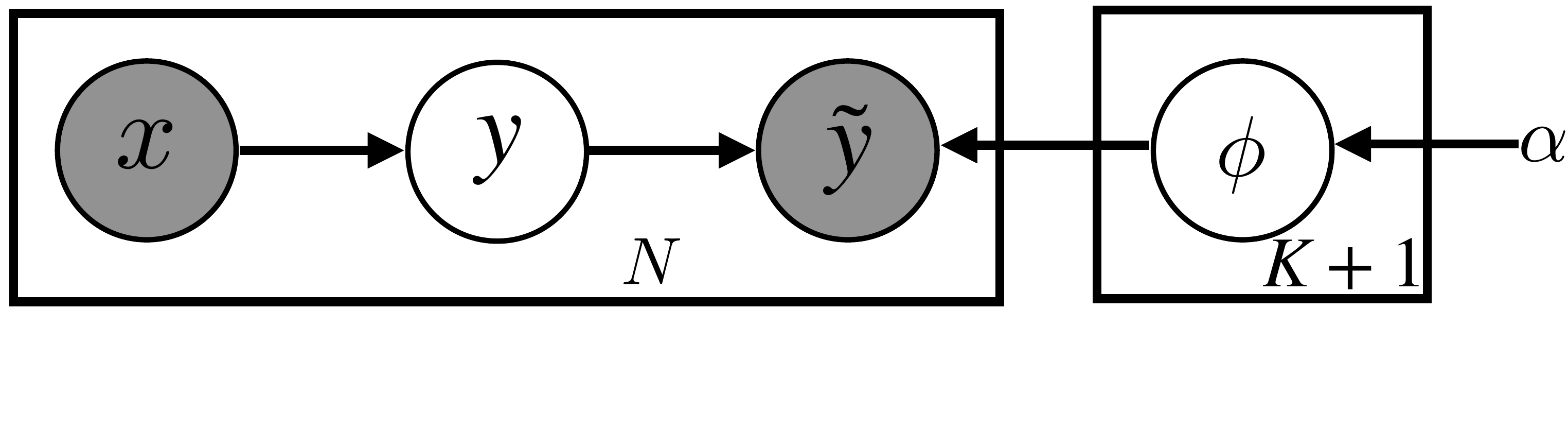}
}\label{lccn:ext1}
\subfigure[LCCN+]{
\includegraphics[width=0.3\textwidth]{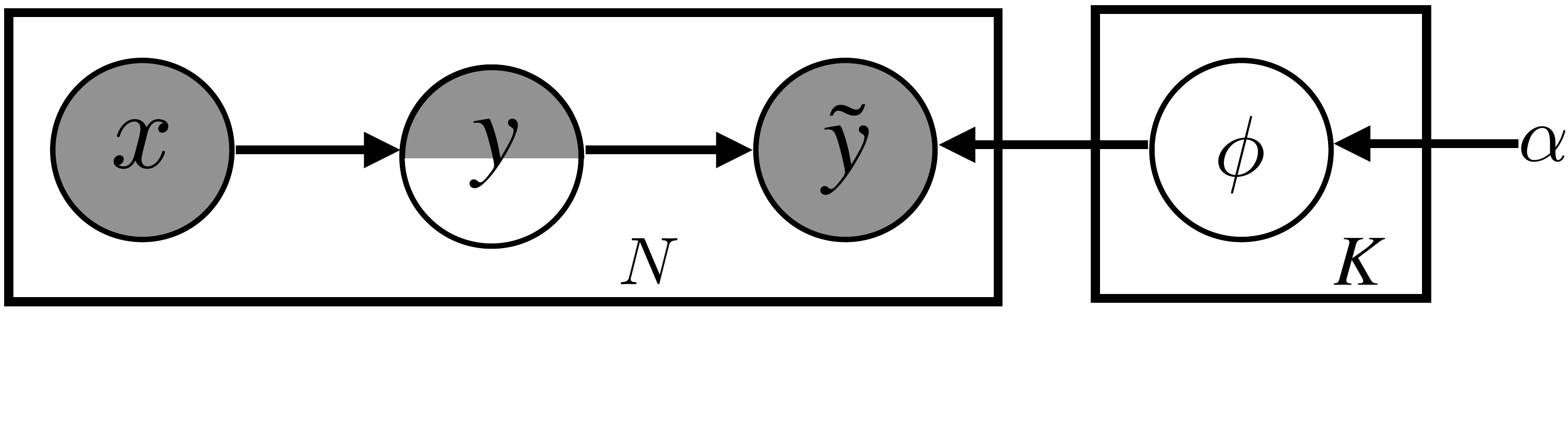}
}\label{lccn:ext2}
\subfigure[DivideLCCN]{
\includegraphics[width=0.3\textwidth]{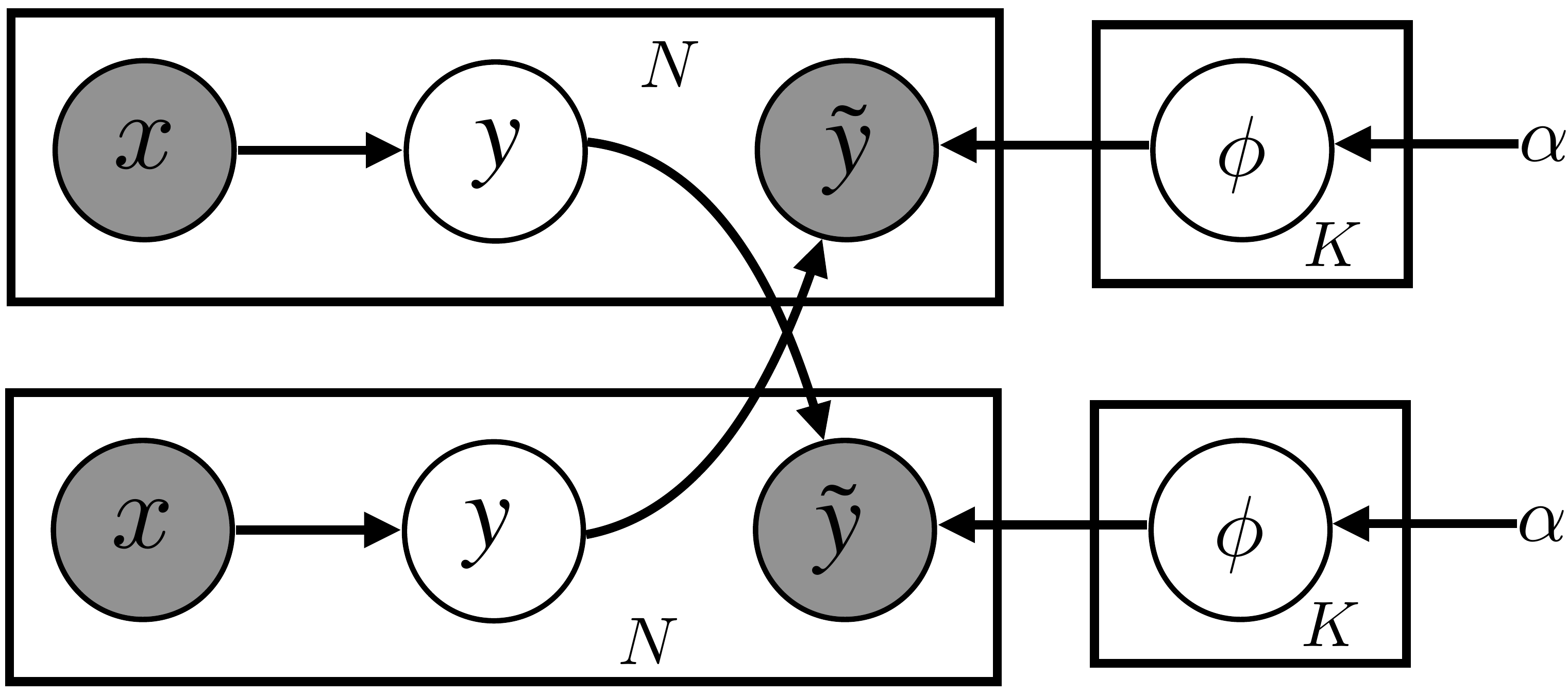}
}\label{lccn:ext3}
\caption{Some variants of the Latent Class-Conditional Noise model. (a) LCCN$^*$ is for the open-set noisy label setting, where compared to LCCN, $\phi\in\mathbb{R}^{(K+1)\times K}$ introduces the extra dimension that collapses the abnormal categories beyond the pre-defined classes. (b) LCCN+ considers the case that the true label $y$, \textit{i.e.,} groundtruth, is partially observed. We use $y$ to train the classifier directly when it is available, otherwise infer it before the training. (c) DivideLCCN extends LCCN to the dual-model structure, which follows the empirical design~\cite{Han2018CoteachingRT,li2019dividemix} for the cross-model training.}\label{fig:outlier_semi}
\end{figure*}

It is an effective way to improve the model performance by learning from the large-scale noisy dataset with a small set of clean samples. Many works~\cite{xiao2015learning,veit2017learning,patrini2017making,pmlr-v80-ren18a,Guo_2018_ECCV} have utilized such a semi-supervised paradigm to calibrate the classifier and achieve a promising result. For LCCN, it is naturally compatible with this, where we can directly use the clean annotations if they are available instead of the labels inferred from Algorithm~\ref{alg:ags}, and simultaneously ignore their statistics to learn the noise transition in Algorithm~\ref{alg:lm}.  
We simply illustrate the corresponding model in Fig.~\ref{fig:outlier_semi}(b).
In a broad sense, clean labels can be as accurate as a fine-grained category index for each sample, or as weak as a coarse hint that tells outlier~\cite{liang2017enhancing}, complementary labels~\cite{ishida2017learning} or not. In the former case, a categorical cross entropy loss can be applied to the classifier; while in the latter case, the binary
cross entropy loss that defines outliers \textit{vs.} non-outliers or the complementary loss should be applied. Considering the coarse labels are greatly different from our scope, we leave the latter case in the future and only validate the former semi-supervised paradigm in this paper.

\subsubsection{Extension for Cross-model Training} 
Cross-model training is a promising method to boost the model performance and avoid the model collapse by means of predictions from other parties~\cite{wu2021cooperative}. It chooses the more confident samples from other networks instead of itself, \textit{e.g.,} the small-loss samples in Co-teaching~\cite{Han2018CoteachingRT} and the refined samples in DivideMix~\cite{li2019dividemix}, to train the classifier. We follow the similar methodology to construct the variant of LCCN. Concretely, we introduce an DivideLCCN model in Fig~\ref{fig:outlier_semi}(c), which exchanges the guessing knowledge about the latent label $y$ on the basis of LCCN. The only change to the optimization is that we would not use the label directly inferred from Eq.~\eqref{eq:step3} but from the refined label like DivideMix. That is, a mixture model about the prediction of $P(y_n|x_n)$ is used to divide the sample into two parts, and then a co-guessing procedure including label augmentation, sharpening and re-mixing are used to refine the label for training. More details can be referred to the Line 4-20 in Algorithm 1 of~\cite{li2019dividemix}. We plot the graphical notation of DivideLCCN in Fig.~\ref{fig:outlier_semi}(3), and will demonstrate our DivideLCCN achieves the state-of-the-art results on a range of datasets in the experimental part.

\section{Experiments}
In this section, we compare our LCCN and its variants with several state-of-the-art methods on both the simulated noisy datasets and the real-world noisy datasets. 

\subsection{Experimental Setup}
\subsubsection{Datasets}
The experiments are conducted on CIFAR-10, CIFAR-100, Clothing1M and WebVision datasets. CIFAR-10 and CIFAR-100~\cite{krizhevsky2009learning} respectively consist of 60,000 32x32 color images from 10 and 100 classes, and both of them contain 50,000 training samples and 10,000 test samples. We inject the symmetric noise and the asymmetric noise to disturb the labels of training samples, which forms the datasets of the close-set label noise. Concretely, in the case of the \emph{asymmetric} noise, we set a probability $r$ to disturb the label to its similar class on CIFAR-10, i.e., truck $\rightarrow$ automobile, bird $\rightarrow$ airplane, deer $\rightarrow$ horse, cat $\rightarrow$ dog. For CIFAR-100, a similar $r$ is set but the label flip only happens in each super-class. The label is randomly disturbed into the next class circularly within the super-classes. In the case of the \emph{symmetric} noise, we follow the noise setting in~\cite{li2019dividemix} to conduct the different ratios of noise. For the experiments that consider the \emph{open-set} noisy labels, we randomly select 10,000 samples from the original datasets and shuffle the order of the pixel values as the samples out of the pre-defined categories. In the semi-supervised learning, we use the clean labels of the 5,000 clean samples and the 500 outlier samples during training.

Regarding Clothing1M~\cite{xiao2015learning}, it has 1 million noisy clothes samples collected from the shopping websites. The authors in~\cite{xiao2015learning} pre-defined 14 categories and assigned the clothes images with the labels that are extracted from the surrounding text provided by sellers, which is thus very noisy. According to~\cite{xiao2015learning}, only about $61.54\%$ labels are reliable. Besides, this dataset contains 50k, 14k and 10k clean samples respectively for the auxiliary training, validation and test. WebVision~\cite{li2017webvision} is a more challenging noisy dataset, which contains more than 2.4 million images. It is crawled from the Internet by using the 1,000 concepts of ILSVRC~\cite{imagenet_cvpr09} as queries. In addition, a clean validation set which contains 50,000 annotated images, are provided to boost and validate the proposed models in diverse applications. We follow the setting of~\cite{li2019dividemix,jiang2020beyond} to use the samples of the first 50 among the total 1000 categories, and evaluate the models on the test set of Webvision and the validation set of ImageNet~\cite{imagenet_cvpr09}.

\subsubsection{Baselines} For the simulated experiments, we compare LCCN and its variants with the classifier that is directly trained on the dataset (termed as CE), the method Bootstrapping proposed in~\cite{reed2014training}, \revision{PES~\cite{bai2021understanding}}, the noise-transition-based method Forward~\cite{patrini2017making} and the method that fine-tunes the transition S-adaptation~\cite{goldberger2016training}, \revision{T-revision~\cite{xia2019t_revision} and brNet~\cite{bootkrajang2022towards}}. Note that, we choose the hard mode for Bootstrapping, since it is empirically better than the soft mode. For the experiments on real-world datasets, we include Joint Optimization~\cite{tanaka2018joint} that leverages the auxiliary noisy label distribution, and the state-of-the-art result Forward+~\cite{patrini2017making} that fine-tunes on clean samples. The Co-teaching+~\cite{yu2019does}, Mixup~\cite{zhang2017mixup}, P-correction~\cite{yi2019probabilistic}, M-correction~\cite{arazo2019unsupervised}, Meta-Learning~\cite{li2019learning}, MentorMix~\cite{jiang2020beyond} and DivideMix~\cite{li2019dividemix} are also considered and keep their original settings, if the results are available in their paper. Note that, we have not required all methods to follow a universal training schedule, since some approaches do not open their codes and the reproduced implementation is not easy to match their original paper. In this case, we will specially follow their original training schedules and mark them in the tables or captions for the fair comparison.

\begin{table*} [t]
\centering
\caption{The average accuracy ($\%$) over 5 trials on CIFAR-10 and CIFAR-100 with different levels of the symmetric noise. OTS means following \underline{O}ur \underline{T}raining \underline{S}chedule and TTS means following \underline{T}heir \underline{T}raining \underline{S}chedules in the original paper. DivideLCCN follows the schedule of DivideMix.} 
{ 
\scalebox{1.0}{
    \begin{tabular}{ c | c | c | c  c  c  c   | c  c  c  c   }
    \hline
    \multicolumn{3}{c|}{Dataset} & \multicolumn{4}{c|}{CIFAR-10} & \multicolumn{4}{c}{CIFAR-100} \\ \hline
   & $\#$ & Method \textbackslash~Noise Ratio & 0.2 & 0.5 & 0.8 & 0.9 & 0.2 & 0.5 & 0.8 & 0.9  \\ \hline \hline
\multirow{5}{*}{OTS}  & 1 & CE & 86.7$\pm$0.5    & 79.3$\pm$0.6  & 62.9$\pm$0.9    & 42.8$\pm$1.5 & 62.0$\pm$0.6   & 46.5$\pm$0.5 & 19.8$\pm$1.1    & 10.0$\pm$0.5 \\ \cline{2-11}
 & 2 & Bootstrapping & 86.9$\pm$0.4    & 79.8$\pm$0.5   & 63.5$\pm$1.0    & 42.6$\pm$1.4 & 62.2$\pm$0.6    & 46.6$\pm$0.6   & 19.9$\pm$0.8    & 10.1$\pm$0.4 \\ \cline{2-11}
& 3 & Forward & 86.9$\pm$0.6    & 79.8$\pm$0.7   & 63.4$\pm$0.6    & 42.7$\pm$1.3 & 61.4$\pm$0.4    & 46.8$\pm$1.0   & 20.0$\pm$0.8    & 10.1$\pm$0.1 \\ \cline{2-11}
& 4 & S-adaptation     & 87.2$\pm$0.6    & 83.4$\pm$0.7    & 64.1$\pm$1.6    & 43.2$\pm$1.4  & 62.7$\pm$0.6 & 48.5$\pm$0.9 & 19.2$\pm$1.5 & 9.1$\pm$0.9    \\  \cline{2-11}

& \revision{5} & \revision{brNet} &\revision{87.1$\pm$0.6} &\revision{83.4$\pm$0.6} &\revision{65.1$\pm$1.4} &\revision{42.3$\pm$1.2} &\revision{63.2$\pm$0.2} &\revision{48.7$\pm$1.0} &\revision{19.9$\pm$1.1} &\revision{8.9$\pm$0.5}   \\ \cline{2-11}

& \revision{6} & \revision{T-revision} & \revision{87.5$\pm$0.8} & \revision{81.9$\pm$0.4} & \revision{48.6$\pm$0.9} & \revision{28.6$\pm$1.4} & \revision{62.0$\pm$0.3} & \revision{46.0$\pm$0.7} & \revision{7.5$\pm$0.2} & \revision{2.0$\pm$0.3}   \\ \cline{2-11}

& \revision{7} & \revision{PES} & \revision{87.9$\pm$0.5} & \revision{84.1$\pm$0.3} & \revision{54.1$\pm$0.7} & \revision{18.0$\pm$0.5} &\revision{62.5$\pm$0.5} &\revision{50.1$\pm$1.3} &\revision{18.0$\pm$0.9} &\revision{8.1$\pm$0.6}   \\ \cline{2-11}

& 8 & LCCN     & \textbf{\underline{88.7}$\pm$0.4}    & \textbf{\underline{87.3}$\pm$0.5} &  \textbf{\underline{67.5}$\pm$0.8}    & \textbf{43.8$\pm$1.2}  & \textbf{\underline{64.6}$\pm$0.2} & \textbf{\underline{52.3}$\pm$0.6} & \textbf{20.2$\pm$0.7} & \textbf{10.4$\pm$0.2}    \\\hline \hline
\multirow{9}{*}{TTS}  
& 9 & Co-teaching+ & 89.7$\pm$0.2    & 85.6$\pm$0.5   & 67.4$\pm$0.5    & 47.9$\pm$1.9 & 65.4$\pm$0.1    & 51.9$\pm$0.2   & 27.7$\pm$0.5    & 13.9$\pm$0.9 \\ \cline{2-11}
& 10 & Mix-up & 95.5$\pm$0.6    & 87.2$\pm$0.2  & 71.9$\pm$0.4    & 52.4$\pm$0.8 & 68.1$\pm$0.9    & 57.4$\pm$0.7   & 30.5$\pm$0.8    & 14.0$\pm$0.7 \\ \cline{2-11}
& 11 & P-correction & 92.3$\pm$0.3    & 89.1$\pm$0.4   & 77.6$\pm$0.6    & 58.7$\pm$1.0   & 69.4$\pm$0.4    & 57.5$\pm$0.6   & 31.3$\pm$0.4    & 15.6$\pm$0.8 \\ \cline{2-11}
& 12 & Meta-Learning & 93.1$\pm$0.1    & 89.1$\pm$0.6   & 77.2$\pm$0.5 & 59.2$\pm$1.4 & 68.4$\pm$0.7    & 59.2$\pm$0.5   & 42.1$\pm$0.5    & 19.7$\pm$0.8 \\ \cline{2-11}
& 13 & M-correction & 94.1$\pm$0.4    & 92.3$\pm$0.3   & 86.8$\pm$0.6    & 69.0$\pm$1.2   & 74.0 $\pm$0.3   & 66.2$\pm$0.7   & 48.3$\pm$0.5    & 24.6$\pm$1.1 \\ \cline{2-11}
& 14 & MentorMix & 95.2$\pm$0.2   & 92.9$\pm$0.2  & 81.2$\pm$0.6  & 50.7$\pm$2.4   & \textbf{78.6$\pm$0.5} & 68.0$\pm$0.5 & 41.1$\pm$0.8   & 15.5$\pm$1.6 \\ \cline{2-11}
& 15 & DivideMix & 96.0$\pm$0.1   & 94.6$\pm$0.1   & \textbf{93.1$\pm$0.4}  & 75.7$\pm$1.1   & 77.4$\pm$0.2    & 74.1$\pm$0.6   & 57.7$\pm$0.4  & 30.8$\pm$0.8 \\ \cline{2-11}
& 16 & DivideLCCN & \textbf{\underline{96.4}$\pm$0.2}    & \textbf{\underline{95.3}$\pm$0.1}   & 93.0$\pm$0.5   & \textbf{\underline{79.2}$\pm$1.6}   & 78.5$\pm$0.3   & \textbf{74.5$\pm$0.5}   & \textbf{\underline{58.6}$\pm$0.4}   & \textbf{31.7$\pm$0.8} \\ \hline
 \end{tabular}}}
\label{tab:cifar_sym}
\end{table*}

\begin{table*} [t]
\centering
\caption{The average accuracy ($\%$) over 5 trials on CIFAR-10 and CIFAR-100 with different levels of the asymmetric noise. The methods of $\#1-\#5$ follow OTS and the other methods \textit{i.e.,} $\#6-\#8$, follow TTS.} 
{ 
\scalebox{0.9}{
    \begin{tabular}{ c | c | c  c  c  c  c | c  c  c  c  c }
    \hline
    \multicolumn{2}{c|}{Dataset} & \multicolumn{5}{|c|}{CIFAR-10} & \multicolumn{5}{|c}{CIFAR-100} \\ \hline
   $\#$ & Method \textbackslash~Noise Ratio & 0.1 & 0.3 & 0.5 & 0.7 & 0.9 & 0.1 & 0.2 & 0.3 & 0.4 & 0.5 \\ \hline\hline
  1 & CE & 90.0$\pm$0.6 & 88.2$\pm$0.5 & 76.7$\pm$1.3 & 58.9$\pm$1.9 & 56.8$\pm$1.1
  & 66.1$\pm$0.5 & 64.1$\pm$0.5 & 60.1$\pm$0.5 & 51.6$\pm$0.9 & 33.3$\pm$1.8 \\ \hline
  2 & Bootstrapping & 90.8$\pm$0.5 & 88.0$\pm$0.6 & 76.1$\pm$1.1 & 57.2$\pm$1.3 & 56.8$\pm$1.5
  & 66.5$\pm$0.3 & 64.5$\pm$0.4 & 63.2$\pm$0.3 & 55.4$\pm$0.7 & \textbf{34.5$\pm$1.1} \\ \hline
 3 & Forward & 90.8$\pm$0.5 & 88.9$\pm$0.9 & 82.6$\pm$1.2 & 67.2$\pm$1.8 & 57.6$\pm$1.4
 & 65.3$\pm$0.4 & 62.8$\pm$0.5 & 61.4$\pm$0.7 & 52.4$\pm$0.9 & 33.6$\pm$1.0 \\ \hline
 4 & S-adaptation & 91.1$\pm$0.2 & 88.6$\pm$0.4 & 86.8$\pm$1.2 & 72.8$\pm$1.6 & 61.1$\pm$0.8
 & 65.5$\pm$0.2 & 64.3$\pm$0.3 & 62.2$\pm$0.2 & 52.6$\pm$0.8  & 30.2$\pm$1.2 \\ \hline
 \revision{5} & \revision{brNet} &\revision{90.5$\pm$0.2} &\revision{89.0$\pm$0.4} &\revision{85.8$\pm$1.1} &\revision{73.2$\pm$1.6}  &\revision{59.1$\pm$1.1}   
 &\revision{65.2$\pm$0.2} &\revision{64.9$\pm$0.4} &\revision{62.4$\pm$0.3} &\revision{54.2$\pm$0.6} &\revision{31.2$\pm$1.9}  \\ \hline
 
 \revision{6} & \revision{T-revision} &\revision{89.8$\pm$0.3} &\revision{88.0$\pm$0.5} &\revision{86.50$\pm$0.8} &\revision{77.2$\pm$1.5}  &\revision{\textbf{\underline{66.6}$\pm$0.9}}  &\revision{67.14$\pm$0.3} &\revision{60.6$\pm$0.3} &\revision{54.3$\pm$0.6} &\revision{45.2$\pm$0.5} &\revision{33.2$\pm$1.6}  \\ \hline
 
 \revision{7} & \revision{PES} &\revision{89.6$\pm$0.6} &\revision{88.4$\pm$0.5} &\revision{87.3$\pm$1.2} &\revision{58.0$\pm$1.5}  &\revision{57.8$\pm$1.7}  
 &\revision{67.0$\pm$0.4} &\revision{66.4$\pm$0.5} &\revision{62.1$\pm$0.5} &\revision{55.9$\pm$0.9} &\revision{31.5$\pm$1.4} \\ \hline
 
 8 & LCCN & \textbf{91.4$\pm$0.4} & \textbf{89.4$\pm$0.4} & \textbf{\underline{88.6}$\pm$0.7} & \textbf{\underline{79.6}$\pm$1.2} & 64.7$\pm$0.8 
 & \textbf{\underline{67.7}$\pm$0.1} & \textbf{\underline{67.8}$\pm$0.1} & \textbf{\underline{66.9}$\pm$0.4} & \textbf{\underline{65.5}$\pm$0.5} & 33.7$\pm$0.6 \\ \hline \hline
 9 & MentorMix & 86.3$\pm$0.4   & 87.5$\pm$0.4   & 56.4$\pm$0.7   & 55.3$\pm$1.9   & 55.8$\pm$1.8
 & 66.2$\pm$0.6  & 66.4$\pm$0.5  & 63.7$\pm$0.7   & 55.9$\pm$0.4   & 30.9$\pm$1.1 \\ \hline
 10 & DivideMix & 95.4$\pm$0.2 & 92.1$\pm$0.5 & 90.5$\pm$0.4 & 75.1$\pm$1.5 & 61.3$\pm$1.2 
 & 77.4$\pm$0.5  & 75.9$\pm$0.4  & 71.5$\pm$0.5   & 58.3$\pm$0.6   & 41.4$\pm$0.5 \\ \hline
 11 & DivideLCCN & \textbf{95.7$\pm$0.5}   & \textbf{\underline{95.6}$\pm$0.4}   & \textbf{\underline{95.0}$\pm$0.2}   & \textbf{\underline{81.5}$\pm$1.7}   &  \textbf{\underline{75.8}$\pm$1.1} 
 & \textbf{\underline{79.1}$\pm$0.4}   & \textbf{\underline{77.6}$\pm$0.4}   & \textbf{\underline{76.5}$\pm$0.3}   & \textbf{\underline{73.6}$\pm$0.7}   &  \textbf{\underline{42.8}$\pm$0.6} \\ \hline
 \end{tabular}}
}
\label{tab:cifar}
\end{table*}

\subsubsection{Implementation}
For CIFAR-10 and CIFAR-100, the PreAct ResNet-32~\cite{he2016deep} is adopted as the classifier. The image is augmented by horizontal random flip and 32$\times$32 random crops after padding with four pixels. Then, the per-image standardization is used. Regarding the optimizer, we utilize SGD with a momentum of 0.9 and a weight decay of $0.0005$. The batch size is set to 128. The training runs 120 epochs in total and is divided into three phases at the $40$th epoch and the $80$th epoch. In these three phases, we respectively set the learning rate as 0.5, 0.1 and 0.01. 
Following the benchmark in~\cite{patrini2017making}, we use CE to initialize the classifier in other baselines and LCCN. For S-adaptation, Eq.~\eqref{eq:cold-start} is used to warm-up the transition parameters in the first 80 epochs. Similarly, we apply this to LCCN for the first 20,000 steps on CIFAR-10. However, on CIFAR-100, we use the diagonal matrix in the warming-up as Eq.~\eqref{eq:cold-start} will induce the high sampling variance for large number of categories and slowly converge. 

\begin{table*} [t!]
\centering
\caption{The average accuracy ($\%$) over 5 trials on CIFAR-10 and CIFAR-100 with different levels of the open-set noise. All methods follow OTS.} 
{ 
\scalebox{0.9}{
    \begin{tabular}{ c | c | c  c  c  c  c | c  c  c  c  c }
    \hline
    \multicolumn{2}{c|}{Dataset} & \multicolumn{5}{|c|}{CIFAR-10} & \multicolumn{5}{|c}{CIFAR-100} \\ \hline
   $\#$ & Method \textbackslash~Noise Ratio & 0.1 & 0.3 & 0.5 & 0.7 & 0.9 & 0.1 & 0.2 & 0.3 & 0.4 & 0.5 \\ \hline \hline
  1 & CE & 89.2$\pm$0.7 & 87.0$\pm$0.4 & 74.5$\pm$1.6 & 62.5$\pm$1.5 & 57.2$\pm$1.7
  & 62.9$\pm$0.4 & 59.8$\pm$0.6 & 55.1$\pm$0.5 & 45.5$\pm$0.9 & 31.5$\pm$1.6 \\ \hline 
  2 & Bootstrapping & 89.9$\pm$0.6 & 84.6$\pm$0.5 & 74.6$\pm$1.3 & 54.6$\pm$1.5 & 55.7$\pm$1.5
  & 63.9$\pm$0.4 & 60.8$\pm$0.4 & 59.5$\pm$0.5 & 40.3$\pm$0.7 & 31.9$\pm$1.3 \\ \hline
 3 & Forward & 88.9$\pm$0.4 & 85.1$\pm$0.7 & 78.4$\pm$1.2 & 58.0$\pm$1.7 & 56.6$\pm$1.4
 & 63.8$\pm$0.5 & 62.4$\pm$0.4 & 61.8$\pm$0.7 & 51.6$\pm$0.8 & 35.9$\pm$1.2 \\ \hline
 4 & S-adaptation & 88.7$\pm$0.4 & 87.4$\pm$0.5 & 61.1$\pm$1.3 & 57.3$\pm$1.6 & 56.8$\pm$1.2
 & 63.6$\pm$0.5 & 61.7$\pm$0.4 & 60.3$\pm$0.5 & 53.4$\pm$0.7  & 32.3$\pm$1.2 \\ \hline
 5 & LCCN & 88.6$\pm$0.4 & 88.0$\pm$0.6 & 82.3$\pm$0.9 & 69.4$\pm$1.1 & 55.2$\pm$0.9
 & 64.1$\pm$0.1 & 63.0$\pm$0.1 & 61.7$\pm$0.4 & 60.4$\pm$0.5 & 33.7$\pm$0.6 \\ \hline 
 6 & LCCN* & 89.4$\pm$0.5 & 88.6$\pm$0.3 & 84.3$\pm$0.8 & \cellcolor{black!25}72.4$\pm$1.0 & \cellcolor{black!25}56.5$\pm$1.2
 & 64.7$\pm$0.2 & 63.0$\pm$0.3 & 62.7$\pm$0.2 & 62.1$\pm$0.4 & \cellcolor{black!25}32.4$\pm$0.7 \\ \hline  
 7 & LCCN+ & 90.2$\pm$0.1 & 88.7$\pm$0.1 & 88.5$\pm$0.3 & \cellcolor{black!25}87.5$\pm$0.3 & \cellcolor{black!25}86.2$\pm$0.5
 & 65.6$\pm$0.2 & 64.3$\pm$0.2 & 63.4$\pm$0.1 & 63.2$\pm$0.3 & \cellcolor{black!25}62.5$\pm$0.4 \\ \hline 
 \revision{8} & \revision{LCCN*+} &\revision{91.4$\pm$0.2}  &\revision{90.5$\pm$0.1}  &\revision{90.1$\pm$0.3}  &\revision{\cellcolor{black!25}\textbf{\underline{89.4}$\pm$0.3}}  &\revision{\cellcolor{black!25}\textbf{\underline{88.7}$\pm$0.2}} 
 &\revision{67.8$\pm$0.1}  &\revision{66.3$\pm$0.2}  &\revision{66.2$\pm$0.3}  &\revision{67.0$\pm$0.2}  &\revision{\cellcolor{black!25}\textbf{\underline{67.5}$\pm$0.2}} \\ \hline 
 9 & DivideLCCN & \textbf{\underline{96.1}$\pm$0.6} & \textbf{\underline{95.2}$\pm$0.4} & \textbf{\underline{94.1}$\pm$0.4} & \cellcolor{black!25}79.2$\pm$1.3 & \cellcolor{black!25}72.5$\pm$1.2 & \textbf{\underline{77.8}$\pm$0.4} & \textbf{\underline{77.3}$\pm$0.5} & \textbf{\underline{76.3}$\pm$0.5} & \textbf{67.1$\pm$0.6} & \cellcolor{black!25}37.2$\pm$0.4 \\ \hline
    \end{tabular}}

}
\label{tab:cifar-outlier}
\end{table*}

\begin{figure*}[!t]
\centering
\centering
\includegraphics[width=0.96\textwidth]{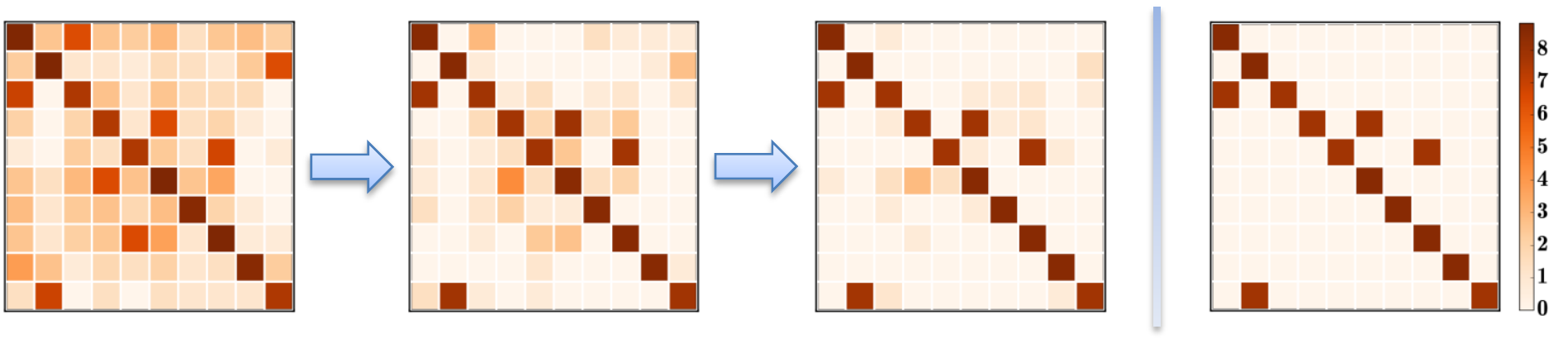}
\caption{The colormap of the confusion matrices on CIFAR-10 with r=0.5. We utilize the log-scale for each element in the confusion matrix for better visualization. The left three maps are respectively learned by LCCN at the beginning, 30,000 step and the end, and the right one is the groundtruth.}
\label{fig:transition}
\end{figure*}

For Clothing1M, the ResNet-50 is applied as the classifier. We resize the short side of their images to 224 and do the random crop of 224$\times$224. The training images are augmented with the random flip, whiteness and saturation. For the optimizer, we adopt SGD with a momentum of 0.9 with a weight decay of $10^{-3}$. The batch size for Clothing1M is set to 32 and we fix the learning rate as 0.001 to run 5 epochs. For the warming-up transition, we both validate the one~\cite{xiao2015learning} from manual annotation and the one estimated by Eq.~\eqref{eq:cold-start} for 40,000 steps. Note that, on the large real-world datasets, due to the strong capacity of ResNet-50, it is easy for LCCN to have the degeneration problem, \textit{i.e.,} the sampled latent label is identical to the noisy label. Thus, we norm Eq.~\eqref{eq:step3} with a power annealed coefficient $\max\{\exp{(-\frac{\text{step}}{\text{max}\_\text{step}}*0.8)},0.5\}$ to introduce the sufficient perturbation in avoid of this issue. On WebVision, the inception-resnet v2 is applied as the classifier, the batch size is set to 64, and the learning rate is initialized with 0.01 and divided by 10 after 40 epochs. 
The similar power-annealed strategy for sampling is applied. \revision{In terms of the evaluation, we consider the average classification accuracy on the clean test dataset. Note that, all methods are run multiple times and we use the underline of results to mark the significance achieved by our method based on the statistical t-test.}

\subsection{Results on CIFAR10 and CIFAR-100}
\subsubsection{Classification}
Table~\ref{tab:cifar_sym} summarizes the model performance under the symmetric noise on CIFAR-10 and CIFAR-100. According to the results in our training schedule \textit{i.e.,} OTS, LCCN ($\#$8) outperforms the noise-transition-based methods ($\#$3-$\#$6) and the bootstrapping way ($\#2$) at all noise levels, \revision{although the recent regularization and completion methods for noise transition like brNet and T-revision also show gains.} Compared to other strong baselines in the perspective of reweighting ($\#$9-$\#$15), LCCN that has not incorporated some learning tricks like mixup~\cite{zhang2017mixup} and label co-refinement~\cite{li2019dividemix}, is thus not on par with them. However, its extended version DivideLCCN ($\#$16) that considers these tricks achieves the new state-of-the-art performance. Specially, we achieve better performance than the previous DivideMix at most of noise levels. \revision{The gain from LCCN to DiviceLCCN indicates that some tricks (especially cross-modal training) in previous methods are important to boost the performance. However, we would like to clarify that DivideLCCN is not to wrap the cross-modal training paradigm as a new contribution, but to demonstrate under the same multi-model setting, the corresponding extension of LCCN can be better than current methods.} Note that, the learning schedules of methods $\#$9-$\#$15 follow their original paper \textit{i.e.,} TTS, since they might be carefully optimized under the corresponding noise settings. \revision{In addition, we present an experimental study to roughly regularize the blow-up of the gradient update for noise transition via gradient clipping. We kindly refer the researchers who are interested in this point to the Appendix C.}

Table~\ref{tab:cifar} presents the results of the methods under asymmetric noise on CIFAR-10 and CIFAR-100. Previous methods have not well validated this setting and reported their performance in the paper like symmetric noise. We hereby enumerate 5 noise rates on each dataset and except baselines $\#$2-$\#$4, we included two typical open-source state-of-the-art methods MentorMix~\cite{jiang2020beyond} and DivideMix~\cite{li2019dividemix} as the baselines. According to the results, LCCN has significant advantages in performance compare to other methods. In particular, in the large noise rates, LCCN even achieves the competitive or better accuracy compare to MentorMix and DivideMix. For example, when $r$=0.7 on CIFAR-10 and $r$=0.4 on CIFAR-100, LCCN reaches $79.6\%$ and $65.5\%$, outperforming the best results in DivideMix, $75.1\%$ and $58.3\%$, by about 5$\%$ and 12$\%$ respectively. Besides, DivideLCCN that integrates some augmentation tricks in DivideMix further improves the performance on basis of LCCN. Actually, the asymmetric noise is a stronger type to compare the model robustness than symmetric noise, since it is a more elaborate attack among specific classes instead of the simple random disturbance to all classes. Note that, regarding $r$=0.5 on CIFAR-100, the way to disturb the labels~\cite{patrini2017making} leads that there is one undesired minimum, since two classes are mixed into one class by equal proportion after injecting noise. In this case, it is hard to say which model can be the best. Thus, as can be seen, LCCN and even MentorMix are not better than Bootstrapping and Forward. Interestingly, DivideLCCN that incorporates the label co-refinement mechanism of DivideMix and LCCN achieves the best performance.

Table~\ref{tab:cifar-outlier} shows the results of LCCN and baselines under the open-set noise on CIFAR-10 and CIFAR-100. Compared to the performance in Table~\ref{tab:cifar}, most of the methods have a slight drop in performance. Nevertheless, LCCN achieved the better performance at $r$=0.3, 0.5, 0.7 on CIFAR-10 and $r$=0.1, 0.2, 0.3 and 0.4 on CIFAR-100 than previous models. Specifically, on CIFAR-10, all the other baselines are even not better than CE, i.e., directly training. LCCN$^*$ that considers the open-set noise achieves a further improvement based on LCCN, and after adding clean data, LCCN+ performs better than LCCN$^*$, since the classifier could be consistently calibrated by the clean data during the training. \revision{When applying the auxiliary clean data in LCCN$^*$ to further calibrate the training, namely, LCCN$^*$+, we can similarly achieve consistent improvement over LCCN+ and LCCN$^*$.} However, in the small rates of label noise, DivideLCCN performs better than both LCCN+ and LCCN$^*$, which shows the advantages of the cross-model training. However, in the extreme noise \textit{i.e.,} $r=0.7,~0.9$ on CIFAR-10 and $r=0.5$ on CIFAR-100, all methods are not better than LCCN+ as marked by the grey color in Table~\ref{tab:cifar-outlier}. This means it is quite useful and even necessary to have a small accurately annotated set for the auxiliary training in the extremely-high noise. In a nutshell, the quantitative analysis of Table~\ref{tab:cifar_sym}, Table~\ref{tab:cifar} and Table~\ref{tab:cifar-outlier} demonstrates the superiority of LCCN and its variants compared to the baselines on synthetic datasets.

\begin{figure}[t!]
\centering
 \includegraphics[width=0.5\textwidth]{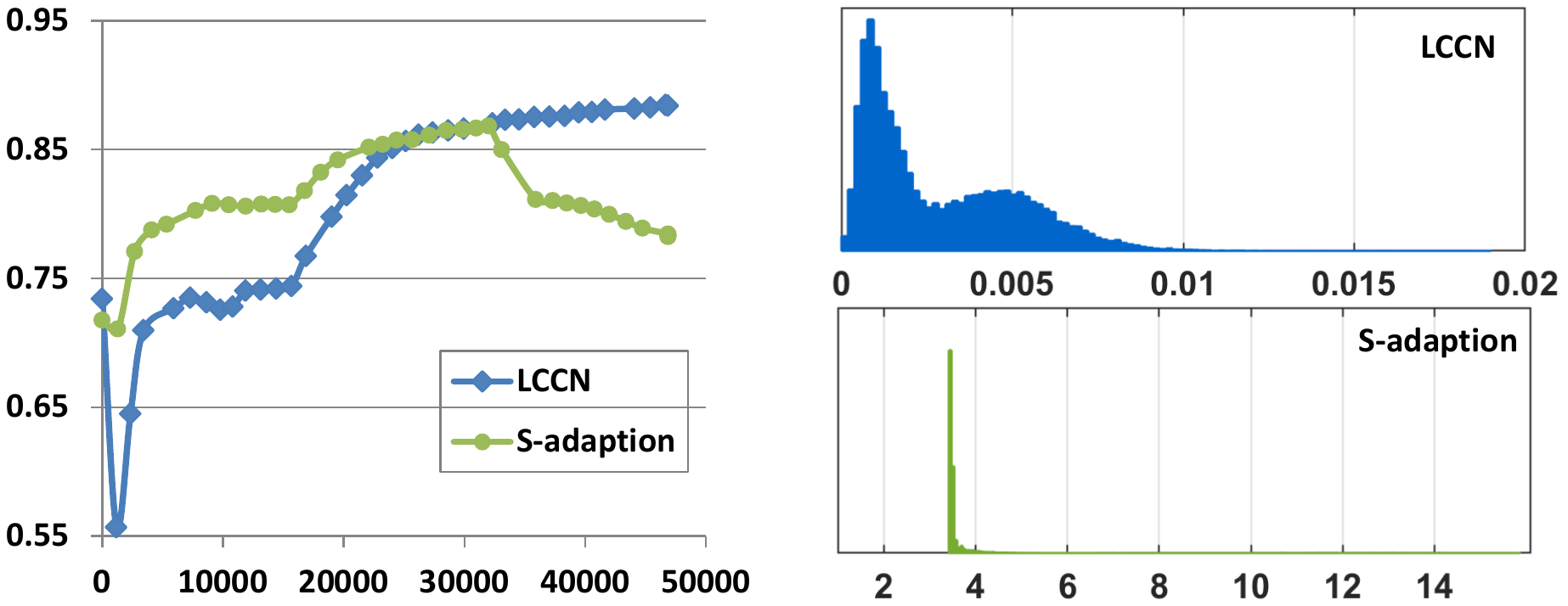}
\caption{The test accuracy of LCCN and S-adaptation in the training on CIFAR-10 with $r$=0.5 (left), and the corresponding histograms for the change of noise transition $\phi$ via a mini-batch of samples (right).}\label{fig:transvar}
\end{figure} 

\begin{figure}[t!]
\centering
\includegraphics[width=0.48\textwidth]{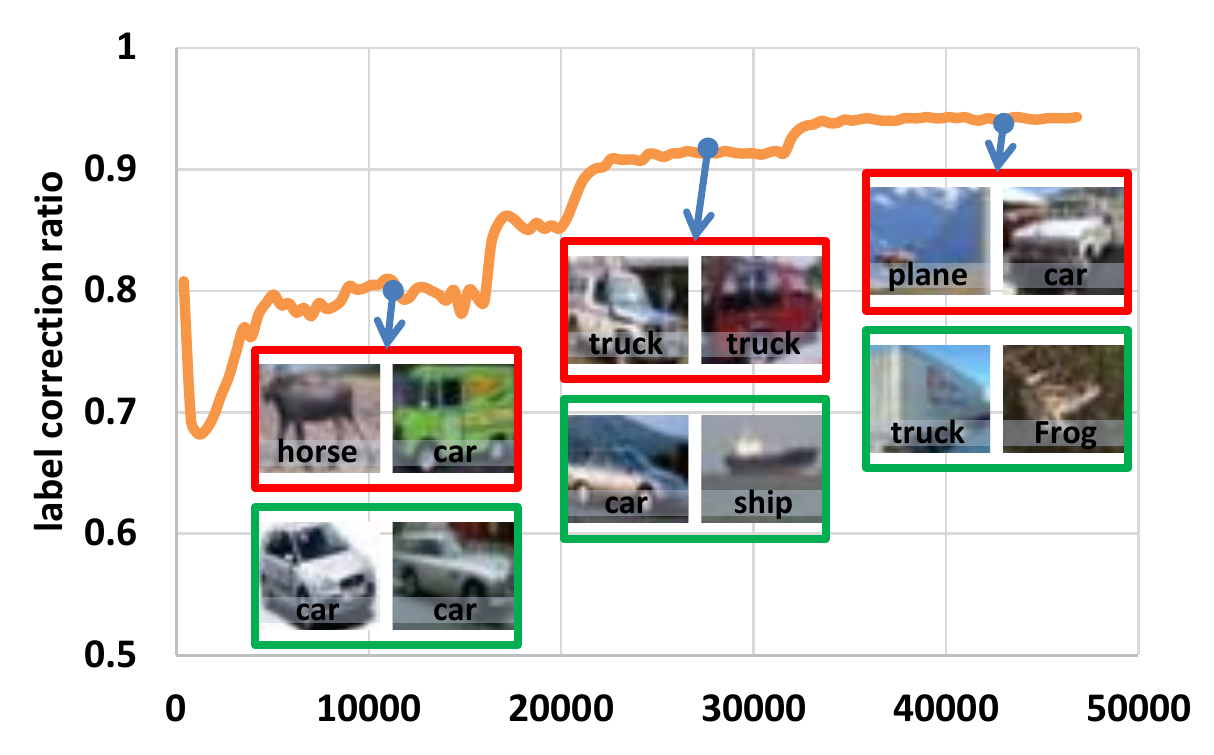}
\caption{The correction ratio during training of LCCN on CIFAR-10 under $r$=0.5 with some negatively corrected samples (the red box) and some positively corrected samples (the green box) in the high probabilities.}
\label{fig:label_trace}
\end{figure}

\subsubsection{Ablation Study}
Fig.~\ref{fig:transition} depicts the colormap of the confusion matrix when training LCCN on CIFAR-10 under the asymmetric noise ($r$=0.5). As can be seen in Fig.~\ref{fig:transition}, the initial confusion matrix does not approach the true transition in many matrix entries. However, along with training, the matrix is gradually corrected \textit{e.g.,} the snapshot at 30,000 step, and at the end of training, it approximately approaches to the given groundtruth. \revision{Here, one might be interested in the case where an oracle noise transition is available in prior. In Appendix D, we conduct more experiments to verify its performance to Forward and LCCN, and discuss its limitation considering other factors for the final performance. More details can be referred to the correponding appendix.} 

To show LCCN can safeguard the noise transition update compared to S-adaptation, we compute the statistics about their update of noise transition on CIFAR-10 at $r$=0.5, and illustrate the histogram of changes in Fig.~\ref{fig:transvar}. Firstly, from the left panel of Fig.~\ref{fig:transvar}, we can see that there is a significant performance drop in the training of S-adaptation. We conjecture that the reason is dramatically updating the noise transition. As shown in the right panel of Fig.~\ref{fig:transvar}, the varying magnitude of $\phi$ in S-adaptation is higher than that of LCCN. One is in a large scale ranging from 0 to 16, while the other one is in a very small scale ranging from 0 to 0.02. This leads to S-adaptation suffering from a high risk of over-tuning towards undesired local minimums in the presence of noise. Conversely, LCCN updates $\phi$ in a small scale when approaching to the minimum, and thus achieve a stable convergence. This quantitative analysis justifies the claim of our Theorem~\ref{theorem:update}. 
Besides, As shown in Fig.~\ref{fig:label_trace}, the ratio of the image with the correct label increases along with the training progress. This reflects LCCN successfully models the class-conditional noise and gradually infer the latent labels. Specifically, by visualizing the mis-corrected examples in the training process, we can find that the classifier at first make mistakes in even some simple samples, while the classifier has the wrong classification on only hard examples finally. These two figures visualize how the dynamic label regression infers the latent label and models the noise.

\begin{figure}[t!]
\centering
\includegraphics[width=0.48\textwidth]{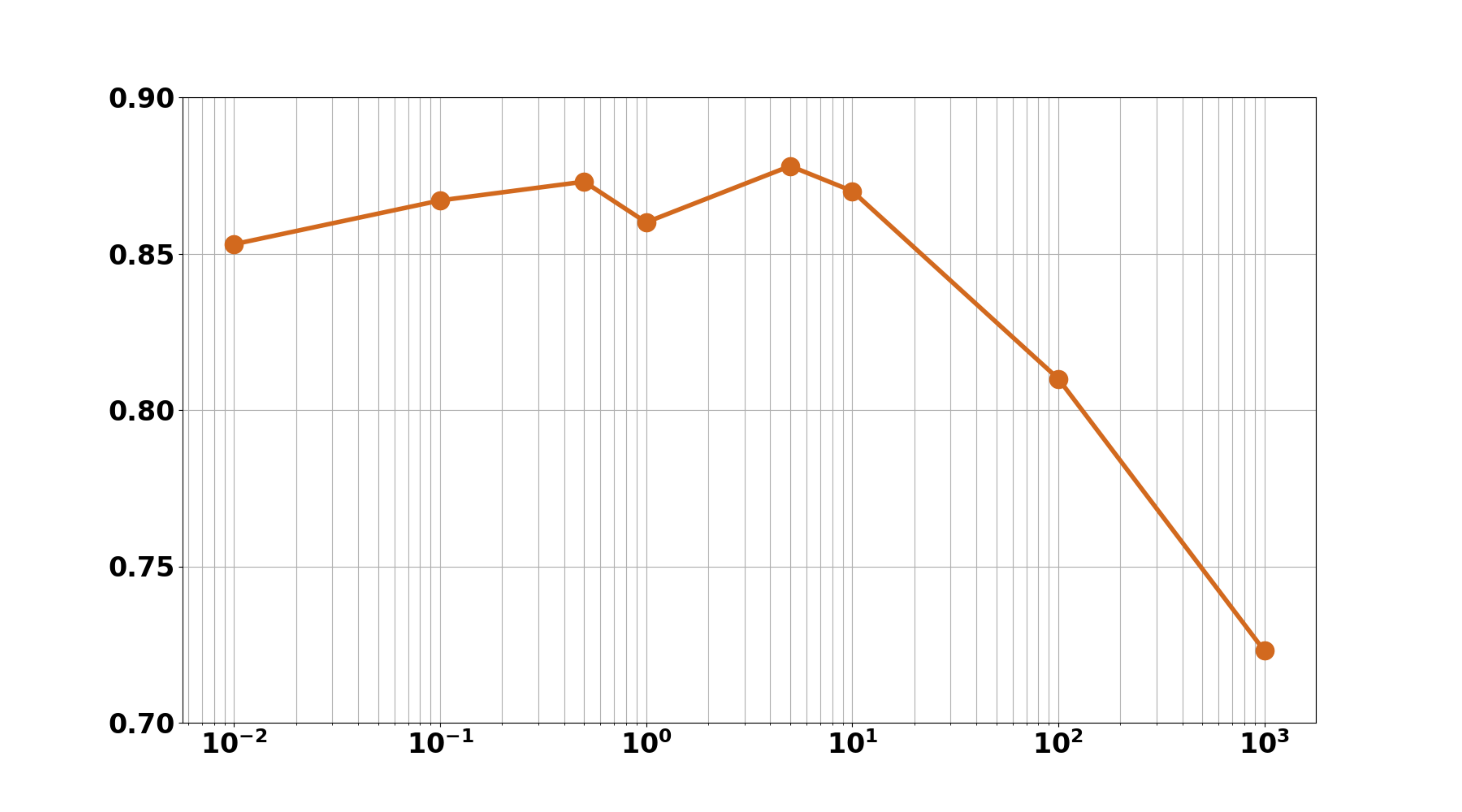}
\caption{Ablation study of the Dirichlet parameter $\alpha$ \textit{w.r.t.} test accuracy. }
\label{fig:alpha}
\end{figure}

\begin{figure*}[!t] 
\centering
\includegraphics[width=0.99\textwidth]{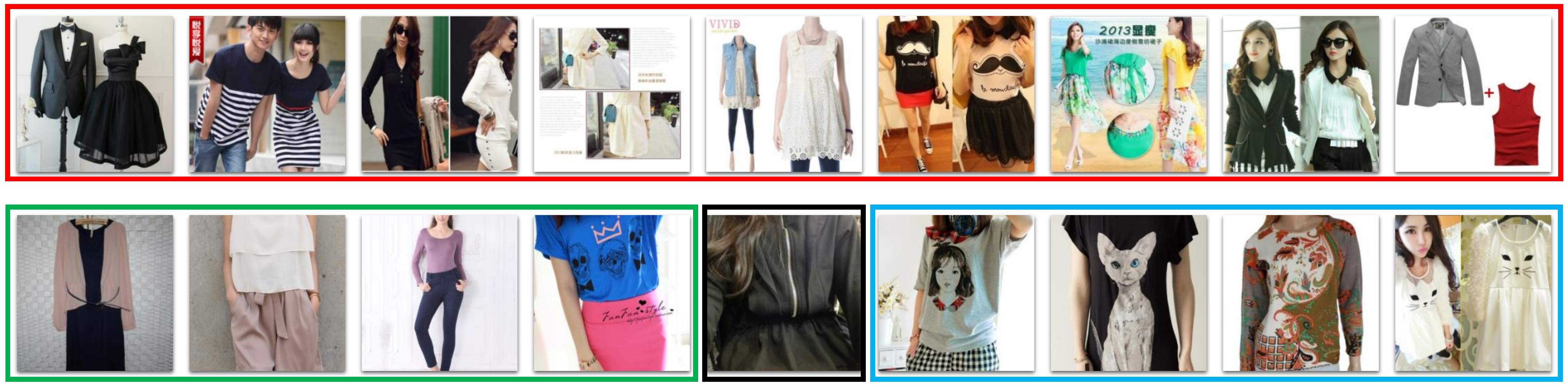}
\caption{Some exemplars that are considered as the open-set label noise by LCCN* in the training set. We intuitively summarize these photos into four categories based on their contents, multiple different objects (RED box), implicit categories (GREEN box), uncertain types (BLACK box) and confusing appearance (BLUE box), which are respectively marked by the color of the surrounded boxes. }\label{fig:clothing1M_outlier}
\end{figure*}

In Eq.~\eqref{eq:step3}, there is $\alpha$ in our model that needs to be set manually. \revision{As the Dirichlet parameter, it affects the concentration of a probability distribution on a simplex. When we apply $Dirichlet(\alpha)$ with a categorical distribution, the posterior naturally follows another Dirichlet distribution, whose Dirichlet parameter is up to the observation statistics smoothed by $\alpha$. Similarly, $\alpha$ in Eq.~\eqref{eq:step3} plays the smoothing role to compute the sampling probability. It is specially important to avoid the deadlock for some minority classes when their allocations becomes very sparse in the Gibbs chain.} In previous experiments, we just set it with a default value 1.0. In this section, we discuss its choice to the model performance. Concretely, we repeat experiments on CIFAR-10 under the asymmetric noise at $r$=0.5 with $\alpha$=0.01, 0.1, 0.5, 1.0, 5.0, 10, 100, 1000 and illustrate the corresponding performance in Fig.~\ref{fig:alpha}. According to the results, we find that the test accuracy at $\alpha$=0.01 is about $85\%$, fluctuates around $88.0\%$ at other $\alpha$ smaller than 10, and decreases significantly to $72.14\%$ when $\alpha$=1000. It indicates as a smoothing parameter in the sparse transition matrix, $\alpha$ should be neither too large nor too small. The small $\alpha$ may reduce the speed of tuning $\phi$ to reach the optimal, in which the entry needs a large transition probability but has a small initial pre-estimation. On the other hand, too large $\alpha$ could also introduce the negative effect to the entry in $\phi$, where there should not be the transition. In this case, the misleading labels will be frequently sampled and mixed into the training due to the excessive smoothing. Correspondingly, the performance is deteriorated. Empirically, we can set $\alpha$ in a proper interval by referring to the sample number of each class to activate the transition when there is no other warming-up available.

\subsection{Results on Clothing1M and WebVision}
Table~\ref{tab:clothing1M} lists the performance of LCCN and baselines on the large-scale Clothing1M dataset. According to the results, we can see that Forward does not show the significant improvement in this dataset, even though they use the annotated noise transition matrix~\cite{xiao2015learning}.  S-adaptation also only improves by 0.5$\%$ compare to Forward. Joint Optimization that trains the classifier with label correction~\cite{tanaka2018joint} achieves better results than the other baselines. Nevertheless, this method requires the additional noisy label distribution to prevent degeneration and is not scalable to the large number of classes~\cite{tanaka2018joint}. Instead, LCCN that contains both the label correction and Bayesian noise modeling, gets the competitive performance 71.63$\%$. With the warming-up of the auxiliary noise transition~\cite{xiao2015learning}, it further achieves the best performance at 73.07$\%$. Even though there is no auxiliary information available, our extension LCCN* still outperforms the current state-of-the-art result. This demonstrates the potential of LCCN in handling real-world noisy datasets. Besides, compare to other recent methods following their training schedules ($\#7-\#12$), our variant DivideLCCN also achieves the competitive performance. Finally, if we have some clean training sample available, we could find that LCCN+ also has the advantages in the semi-supervised learning setting.

\begin{table} 
\caption{The average accuracy over 5 trials on Clothing1M.}
\begin{center}
{
    \begin{tabular}{ c | c | c | c }
     \hline
  \multirow{8}{*}{OTS}  & $\#$ & Method & Accuracy \\
     \hline
     \hline
    & 1 & CE & 68.78$\pm$1.0 \\
     \cline{2-4}
    & 2 & Bootstrapping & 69.05$\pm$0.9\\
     \cline{2-4}
    & 3 & Forward & 69.91$\pm$1.2 \\
     \cline{2-4}
    & 4 & S-adaptation & 70.23$\pm$1.4 \\
     \cline{2-4}
    & 5 & Joint Optimization & 72.33$\pm$0.5 \\
     \cline{2-4}
    & \multirow{3}{*}{6}  & LCCN & 71.78$\pm$0.7 \\
    &  & LCCN* & 72.71$\pm$0.6 \\
    &  & LCCN warmed-up by $\phi$ in~\cite{xiao2015learning}  & \textbf{72.98$\pm$0.4} \\
     \hline \hline
  \multirow{7}{*}{TTS} & 7 & M-correction & 71.05$\pm$0.8 \\
     \cline{2-4}
    & 8 & Meta-Cleaner & 72.34$\pm$0.9 \\
     \cline{2-4}
    & 9 & Meta-Learning & 73.39$\pm$0.7 \\
     \cline{2-4}
    & 10 & P-correction & 73.55$\pm$0.9 \\
     \cline{2-4}
    & 11 & DivideMix & 74.71$\pm$0.8  \\
     \cline{2-4}
    & 12 & ELR+ & 74.75$\pm$1.3 \\
     \cline{2-4}
    & 13 & DivideLCCN & \textbf{74.79$\pm$0.7}  \\
     \hline
     \hline
   \multirow{3}{*}{OTS}  & 14 & CE on the clean data & 75.48$\pm$0.4 \\
     \cline{2-4}
    & 15 & Forward+ & 80.12$\pm$0.5 \\
     \cline{2-4}
    & 16 & LCCN+ & \textbf{\underline{81.34}$\pm$0.5} \\
     \hline
    \end{tabular}
}
\end{center}
\label{tab:clothing1M}
\vspace{-0.5cm}
\end{table}

\begin{table}[t!]
\caption{The learned noise transition on Clothing1M by LCCN$^*$.}
\centering
\renewcommand{\tabcolsep}{0.7mm}
{ \scalebox{1.15}{
\begin{tabular}{c|*{14}{E}|c}
 \multicolumn{1}{c}{class} & \multicolumn{1}{c}{1} 
  & \multicolumn{1}{c}{2} & \multicolumn{1}{c}{3} & \multicolumn{1}{c}{4} 
  & \multicolumn{1}{c}{5} & \multicolumn{1}{c}{6} & \multicolumn{1}{c}{7} & \multicolumn{1}{c}{8} & \multicolumn{1}{c}{9} & \multicolumn{1}{c}{10} & \multicolumn{1}{c}{11} & \multicolumn{1}{c}{12} & \multicolumn{1}{c}{13} & \multicolumn{1}{c}{14} & \multicolumn{1}{c}{} \\ \hhline{~*{14}{|-}|}
  1 & 77 & 0 & 6 & 0 & 0 & 4 & 0 & 0 & 0 & 0 & 0 & 0 & 0 & 0 & {\tiny\bf T-shirt} \\ 
  2 & 2 & 88 & 0 & 3 & 0 & 0 & 0 & 0 & 0 & 0 & 0 & 0 & 0 & 0 & {\tiny\bf shirt}\\ 
  3 & 12 & 4 & 48 & 4 & 7 & 7 & 0 & 0 & 0 & 0 & 0 & 5 & 0 & 0 & {\tiny\bf knitwear}  \\ 
  4 & 9 & 15 & 5 & 51 & 0 & 0 & 0 & 0 & 0 & 0 & 0 & 9 & 0 & 0 & {\tiny\bf chiffon}\\ 
  5 & 11 & 4 & 43 & 0 & 16 & 9 & 0 & 0 & 0 & 0 & 0 & 4 & 0 & 0 & {\tiny\bf sweater} \\ 
  6 & 3 & 0 & 3 & 0 & 0 & 86 & 0 & 0 & 0 & 0 & 0 & 0 & 0 & 0 & {\tiny\bf hoodie}\\ 
  7 & 0 & 0 & 0 & 0 & 0 & 0 & 87 & 3 & 0 & 5 & 0 & 0 & 0 & 0 & {\tiny\bf windbreaker} \\ 
  8 & 0 & 0 & 0 & 0 & 0 & 0 & 0 & 92 & 0 & 0 & 0 & 0 & 0 & 0 & {\tiny\bf jacket}\\ 
  9 & 0 & 0 & 0 & 0 & 0 & 0 & 2 & 0 & 96 & 0 & 0 & 0 & 0 & 0 & {\tiny\bf downcoat}\\ 
  10 & 0 & 0 & 0 & 0 & 0 & 0 & 4 & 2 & 0 & 91 & 0 & 0 & 0 & 0 & {\tiny\bf suit}\\ 
  11 & 0 & 0 & 3 & 0 & 0 & 0 & 0 & 0 & 0 & 0 & 81 & 3 & 0 & 0 & {\tiny\bf shawl}\\ 
  12 & 0 & 0 & 0 & 3 & 0 & 0 & 0 & 0 & 0 & 0 & 0 & 84 & 3 & 0 & {\tiny\bf dress}\\ 
  13 & 0 & 0 & 0 & 3 & 0 & 0 & 0 & 0 & 0 & 0 & 0 & 6 & 76 & 3 & {\tiny\bf vest}\\ 
  14 & 1 & 0 & 0 & 0 & 0 & 0 & 0 & 0 & 0 & 0 & 0 & 0 & 0 & 95 & {\tiny\bf underwear}\\ 
  15 & 12 & 10 & 0 & 7 & 0 & 9 & 6 & 4 & 0 & 7 & 0 & 14 & 6 & 3 & {\tiny\bf outlier}\\ \hhline{~*{14}{|-}|}
\end{tabular} }
}
\label{tab:confusion}
\end{table}

\begin{table*} [!t]
\centering
\caption{The accuracy on (mini)WebVision dataset. Numbers
denote top-1 (top-5) accuracy on the WebVision test set and the ImageNet
validation set.} \label{tab:webvision}
{ 
\scalebox{1.2}{
    \begin{tabular}{  c | c | c | c | c | c }
    \hline
    \multicolumn{2}{c|}{} & \multicolumn{2}{|c|}{WebVision} & \multicolumn{2}{|c}{ILSVRC2012} \\ \hline\hline
   $\#$ & Method & Accuracy$@$1 & Accuracy$@$5 & Accuracy$@$1 & Accuracy$@$5  \\ \hline
  \hline
   
  1 & Forward & 60.98$\pm$0.7    & 82.20$\pm$1.1   & 57.11$\pm$0.6    & 81.93$\pm$0.8 \\ \hline
  2 & Decoupling & 62.73$\pm$0.5   & 85.01$\pm$0.7   & 58.42$\pm$0.5    & 82.53$\pm$0.6 \\ \hline
  3 & D2L & 62.75$\pm$0.4    & 84.31$\pm$0.6   & 57.93$\pm$0.4    & 81.55$\pm$0.5 \\ \hline
  4 & MentorNet & 62.91$\pm$0.5   & 81.70$\pm$0.7  & 57.88$\pm$0.4    & 79.95$\pm$0.7 \\ \hline
  5 & Co-teaching & 63.87$\pm$0.3   & 85.38$\pm$0.6   & 61.54$\pm$0.3    & 84.76$\pm$0.5 \\ \hline
  6 & Iterative-CV & 65.12$\pm$0.6   & 85.05$\pm$0.8   & 61.49$\pm$0.5    & 84.67$\pm$0.8 \\ \hline
  
  7 & LCCN & 72.34$\pm$0.5   & 88.82$\pm$0.7   & 67.48$\pm$0.4    & 88.30$\pm$0.6 \\ \hline
  8 & LCCN$^*$ & 73.05$\pm$0.5   & 90.44$\pm$0.8   & 67.67$\pm$0.5    & 89.84$\pm$0.7 \\ \hline
  9 & ELR+ & 77.63$\pm$0.7    & 91.38$\pm$0.9   & 70.19$\pm$0.6  & 89.43$\pm$1.0 \\ \hline
  10 & MentorMix & 73.58$\pm$0.6    & 91.34$\pm$0.8  & 75.69$\pm$0.7   & 90.43$\pm$0.9 \\ \hline
  11 & DivideMix & 77.64$\pm$0.4    & 91.93$\pm$0.7   & 75.31$\pm$0.3  & 91.20$\pm$0.6 \\ \hline
  12 & DivideLCCN & \textbf{\underline{78.90}$\pm$0.4}   & \textbf{\underline{93.14}$\pm$0.6}   & \textbf{\underline{76.28}$\pm$0.4}   & \textbf{\underline{93.54}$\pm$0.5} \\ \hline
 \end{tabular}}
}
\label{tab:webvision_update}
\end{table*}

In Table~\ref{tab:confusion}, we present the noise transition matrix learned by LCCN*, where only the significant transition probabilities are marked. From this noise transition matrix, we can see that the training samples in some classes are very noisy. For example, \emph{knitwear} and \emph{sweater} are two classes which transit most of labels to other classes. This is because such two classes usually co-occur with other categories like \emph{jacket} or \emph{shawl} in the common dress collocation, which may incur the label transition due to their visual similarity. Besides, in Table~\ref{tab:confusion}, we can find which class contains a lot of outliers from the $15$th transition. Furthermore, to better understand these open-set noise, we give some representative samples in Fig.~\ref{fig:clothing1M_outlier}. According to the image contents, we intuitively summarize the open-set noise into four sub-classes, \textit{i.e.,} multiple different objects, implicit categories, uncertain types and confusing appearance. Based on this visualization, it is usually improper to assign an unique label to these images, since some may contain multiple kinds of clothes. However, clothing1M is naturally considered as the single label classification problem, which may need to be reconsidered in the future. And in some challenging cases, e.g., the images in the blue box in Fig.~\ref{fig:clothing1M_outlier} , the hard example is also considered as the open-set noise by LCCN$^*$, even if the label is correct. Actually, this illustrates the potential drawback of our model in distinguishing hard examples and noisy examples, which requires more explorations in the future.

Table~\ref{tab:webvision_update} presents the performance of LCCN, its variants and a range of baselines on a more challenging noisy dataset WebVision. This task couples two lengthy challenging sub-tasks, \textit{i.e.,} perfectly decoupling the clean samples from the noisy dataset and well fitting the clean samples in this popular benchmark dataset in computer vision. From the current limiting performance of image recognition in ImageNet~\cite{imagenet_cvpr09}, we know either sub-task mentioned above needs careful optimization to achieve a satisfying result in such a challenging scenario. In this paper, we train the model on the noisy WebVision training set as~\cite{li2019dividemix,jiang2020beyond}, and compute both Top-1 and Top-5 accuracies on WebVision and ImageNet in the experiments. According to the results of Table~\ref{tab:webvision_update}, LCCN achieves the better performance compared to the methods of $\#1-\#6$ either in terms of Top-1 accuracy or Top-5 accuracy. Similarly, LCCN$^*$ that considers the open-set label noise in this large-scale dataset have the slight refinement based on LCCN. Compared to the recent advances $\#9-\#11$, LCCN and LCCN$^*$ do not have the advantages in the Top-1 accuracy. However, DivideLCCN that builds upon on recent augmentation tricks and cross-model training with LCCN, achieves the best performance than all previous methods. The results show the promise of LCCN and its variants in the real-world large-scale classification tasks.

\section{Conclusion and Future Work}
In this paper, we present a Latent Class-Conditional Noise model for learning with noisy supervision. A dynamic label regression method is introduced for LCCN to efficiently infer the latent labels, train the classifier and model the class-conditional noise. Theoretical analysis on the model convergence and the essential gap between training and test are also provided to understand the underlying dynamics. Most importantly, we demonstrate that our method safeguards the update of the noise transition unlike the previous arbitrarily tuning by the backpropagation in a neural layer. Finally, we generalize our model to some typical variants, which can handle the open-set label noise, the semi-supervised learning and cross-model training. A range of experiments on the toy and the real-world datasets confirm the superiority of LCCN and its variants. Although we have shown the advantages of LCCN in a range of applications with generalized noisy supervision, other specific settings that consider more complex noise, \textit{e.g.,} instance-dependent noise in the classification or \textit{region-level} noise in the detection and segmentation, are also important in practise, which will be explored in the future works. 

\ifCLASSOPTIONcompsoc
  \section*{Acknowledgments}
\else
  \section*{Acknowledgment}
\fi

This work was supported by STCSM (No. 22511106101, No. 18DZ2270700, No. 21DZ1100100), 111 plan (No. BP0719010), State Key Laboratory of UHD Video and Audio Production and Presentation, NSFC Young Scientists Fund No. 62006202, RGC Early Career Scheme No. 22200720, Guangdong Basic and Applied Basic Research Foundation No. 2022A1515011652, CAAI-Huawei MindSpore Open Fund, and Centre for Frontier AI research.

\appendices
\section{Proof of $\Delta_\mathcal{F}$ regarding to the two quantities}
By definition, the excess risk of the estimated model $\hat{f}_\theta$ is directly bounded by the absolute supremum $\Delta_\mathcal{F}$ as follows,
\begin{align} \label{eq:excess_risk}
\begin{split}
&\mathbf{E}\left[\ell_1(z,\hat{f}_\theta(x))\right]-\mathbf{E}^{(D_N)}\left[\ell_1(z, \hat{f}_\theta(x))\right]\\ 
&\qquad \leq\sup_{f_\theta\in\mathcal{F}}\big|\mathbf{E}\left[\ell_1(z,f_\theta(x))\right]-\mathbf{E}^{(D_N)}\left[\ell_1(z, f_\theta(x))\right]\big|.
\end{split}
\end{align}
Assume the Bayes optimal classifier is $f^*_\theta$. Since $\hat{f}_\theta$ minimizes the empirical loss on $D_N$, we have the following inequality,
$$\mathbf{E}^{(D_N)}\left[\ell_1(z, f^*_\theta(x))\right]-\mathbf{E}^{(D_N)}\left[\ell_1(z, \hat{f}_\theta(x))\right]\geq 0.$$
Then, for the error bound \textit{w.r.t.} the expected risk and the Bayes risk, we can deduce with above inequalities as follows,
\begin{align} \label{eq:error_bound}
\begin{split}
&\mathbf{E}\left[\ell_1(z,\hat{f}_\theta(x))\right]-\mathbf{E}\left[\ell_1(z, f^*_\theta(x))\right]\\ 
&\leq\mathbf{E}\left[\ell_1(z,\hat{f}_\theta(x))\right]-\mathbf{E}^{(D_N)}\left[\ell_1(z, \hat{f}_\theta(x))\right] \\
&\quad -\left( \mathbf{E}\left[\ell_1(z,f^*_\theta(x))\right]-\mathbf{E}^{(D_N)}\left[\ell_1(z, f^*_\theta(x))\right] \right) \\
&\leq 2*\sup_{f_\theta\in\mathcal{F}}\big|\mathbf{E}\left[\ell_1(z,f_\theta(x))\right]-\mathbf{E}^{(D_N)}\left[\ell_1(z, f_\theta(x))\right]\big|.
\end{split}
\end{align}
Eq.~\eqref{eq:excess_risk} and Eq.~\eqref{eq:error_bound} show the supremum of both the excess risk and the error bound are characterized by $\Delta_\mathcal{F}$. Thus, we can universally analyze the corresponding upper bound of $\Delta_\mathcal{F}$ to investigate the generalization performance of LCCN.

\begin{table*}[t!]
    \centering
    \revision{
    \resizebox{1\textwidth}{!}{
    \begin{tabular}{c|ccccc|ccccc}
        \hline
        \multirow{2}{*}{Method$\backslash$Noisy Datasets} & \multicolumn{5}{c|}{CIFAR-10}  & \multicolumn{5}{c}{CIFAR-100} \\
        \cline{2-11} 
        & 0.1 & 0.3 & 0.5 & 0.7 & 0.9 & 0.1 & 0.2 & 0.3 & 0.4 & 0.5 \\
        \hline
        S-adaptation &91.02\% &88.83\% &86.79\% &72.74\% &60.92\% &65.52\% &64.11\% &62.39\% &52.74\% &30.07\% \\
        \hline
        S-adaptation + GC ($\tau = 0.1$) &90.78\% &88.42\% &87.23\% &73.62\% &61.29\% &65.81\% &62.43\% &60.86\% &49.24\% &28.99\% \\
        \hline
        S-adaptation + GC ($\tau = 1.0$) &91.15\% &88.95\% &86.23\% &72.14\% &60.47\% &66.22\% &64.92\% &62.24\% &53.04\% &29.82\% \\
        \hline
        S-adaptation + GC ($\tau = 10.0$) &90.91\% &89.04\% &85.24\% &73.46\% &61.52\% &66.25\% &63.49\% &61.94\% &53.34\% &30.97\% \\
        \hline
        LCCN &91.35\% &89.33\% &88.41\% &79.48\% &64.82\% &67.82\% &67.63\% &66.86\% &65.52\% &33.71\% \\
        \hline
    \end{tabular}}
    \caption{Comparison among S-adaptation, S-adaptation with Gradient Clipping (GC) and LCCN on CIFAR-10 under asymmetric noise ratios 0.1, 0.3, 0.5, 0.7, 0.9 and CIFAR-100 under asymmetric noise ratios 0.1, 0.2, 0.3, 0.4, 0.5. In this table, $\tau$ represents the threshold of gradient clipping.}
    }
    \label{tab:gradient_clipping}
\end{table*}

\begin{table*}[t!]
    \centering
    \revision{
    \resizebox{1\textwidth}{!}{
    \begin{tabular}{c|cccc|cccc}
        \hline
        \multirow{2}{*}{Method$\backslash$Symmetric Noise} & \multicolumn{4}{c|}{CIFAR-10}  & \multicolumn{4}{c}{CIFAR-100} \\
        \cline{2-9} 
        & 0.2 & 0.5 & 0.8 & 0.9 & 0.2 & 0.5 & 0.8 & 0.9 \\
        \hline
        Forward + oracle &88.79\% &81.69\% &55.58\% &31.07\% &58.93\% &48.33\% &17.59\% &9.12\% \\
        \hline
        LCCN + oracle &89.98\% &86.55\% &57.93\% &35.74\% &64.24\% &51.13\% &21.10\% &9.64\% \\
        \hline
        LCCN & 88.6\% & 87.1\% & 67.8\% & 43.7\% & 64.4\% & 52.5\% & 20.3\% & 10.4\% \\
        \hline
        \multirow{2}{*}{Method$\backslash$Asymmetric Noise} & \multicolumn{4}{c|}{CIFAR-10}  & \multicolumn{4}{c}{CIFAR-100} \\
        \cline{2-9} 
        & 0.1 & 0.3 & 0.5 & 0.7 & 0.1 & 0.2 & 0.3 & 0.4 \\
        \hline
        Forward + oracle &91.56\% &87.67\% &76.76\% &57.11\% &69.19\% &67.86\% &66.33\% &65.30\% \\
        \hline
        LCCN + oracle &91.93\% &90.19\% &89.78\% &85.91\% &69.72\% &69.61\% &68.43\% &68.22\% \\
        \hline
        LCCN &91.35\% &89.33\% &88.41\% &79.48\%  &67.82\% &67.63\% &66.86\% &65.52\% \\
        \hline
    \end{tabular}}
    \caption{The performance comparison among Forward with oracle noise transition, LCCN with oracle noise transition and LCCN on CIFAR-10 and CIFAR-100 under different symmetric/asymmetric noise ratios.}
    }
    \label{tab:GT_prior}
\end{table*}

\section{Proof of Theorem~2}
Assume $f^*_\theta$ and $f^\dagger_\theta$ are the underlying groundtruth labeling functions $\mathcal{X}\rightarrow\mathcal{Y}$ of clean test data and data from the Gibbs sampling respectively. Then, the $\Delta_\mathcal{F}$ can be reformulated by applying these notations and further deduced as follows,
\begin{align}\label{eq:theorem2_deduce1}
\begin{split}
& \sup_{f_\theta\in\mathcal{F}}\big|\mathbf{E}\left[\ell_1(z,f_\theta(x))\right]-\mathbf{E}^{(D_N)}\left[\ell_1(z, f_\theta(x))\right]\big| \\
& = \sup_{f_\theta\in\mathcal{F}}\Big|\mathbf{E}\left[\ell_1(f^*_\theta(x),f_\theta(x))\right]-\mathbf{E}^{(D_N)}\left[\ell_1(f^\dagger_\theta(x), f_\theta(x))\right]\Big| \\
& \leq \sup_{f_\theta\in\mathcal{F}}\Big|\mathbf{E}\left[\ell_1(f^*_\theta(x),f_\theta(x))\right]-\mathbf{E}\left[\ell_1(f^\dagger_\theta(x), f_\theta(x))\right]\Big| \\
& + \sup_{f_\theta\in\mathcal{F}}\Big|\mathbf{E}\left[\ell_1(f^\dagger_\theta(x),f_\theta(x))\right]-\mathbf{E}^{(D_N)}\left[\ell_1(f^\dagger_\theta(x), f_\theta(x))\right]\Big| \\
& \leq \underbrace{\sup_{f_\theta\in\mathcal{F}}\Big|\mathbf{E}\left[\ell_1(f^*_\theta(x) - f^\dagger_\theta(x), f_\theta(x))\right]\Big|}_{\Delta} \\
& + \underbrace{\sup_{f_\theta,f'_\theta\in\mathcal{F}}\Big|\mathbf{E}\left[\ell_1(f'_\theta(x),f_\theta(x))\right]-\mathbf{E}^{(D_N)}\left[\ell_1(f'_\theta(x), f_\theta(x))\right]\Big|}_{\Delta_{disc}} 
\end{split}
\end{align}
The second term $\Delta_{disc}$ in the right-hand side of Eq.~\eqref{eq:theorem2_deduce1} is the popular discrepancy distance. It has been demonstrated by the following Rademacher bound
for any probability $\delta>0$,
$$\Delta_{disc}\leq \widehat{\mathcal{R}}(\mathcal{G}) + 3\rho \sqrt{\frac{\ln(\frac{2}{\delta})}{2N}},$$
where $\mathcal{G}$ is defined by the composite functional class $\{x\mapsto \ell_1(f'_\theta(x),f_\theta(x)):f'_\theta,f_\theta\in\mathcal{F}\}$ and $N$ is the sample number. Then, combined with the given Rademacher bound, we finally proof the Theorem~2. i.e., for any probability $\delta$, the generalization bound of LCCN is 
\begin{align}\label{eq:theorem2_deduce2}
\begin{split}
& \Delta_\mathcal{F} \leq \Delta + \widehat{\mathcal{R}}(\mathcal{G}) + 3\rho \sqrt{\frac{\ln(\frac{2}{\delta})}{2N}}.
\end{split}
\end{align}
This bound theoretically points out three important factors to affect our model generalization performance, i.e., domain gap, the function complexity and the sample number.

\revision{
\section{Gradient Clipping in Transition Adaptation}
In the aforementioned section, we have discussed that the instability issue of transition update in previous methods is the bottleneck for transition-based methods with DNNs. One straightforward idea is that we can use some regularization way like gradient clipping to prevent the blow-up of gradient estimation for noise transition in previous methods. To validate the effectiveness of this naive idea, we add a study in this part by applying gradient clipping to S-adaptation. The experimental results are summarized in Table~\ref{tab:gradient_clipping}. From the results, we can find that roughly clipping gradients cannot consistently improve the performance. The potential explanation is that the large gradients might not always mean the confusing signal perturbed by the extreme noise, but also can be the meaningful learning signal from the clean samples (especially clean hard examples). Naively clipping all large gradients might result in the insufficient training without the support of some important hard examples, limiting the improvement of the model performance. We have added these results in the appendix and clarified this point in the experimental parts.
}

\revision{
\section{Learning with the oracle transition}
Although in the class-conditional noise model, we mainly focus on improve an accurate estimation of noise transition, it is lack of sufficient discussion about its extreme. In Table~\ref{tab:GT_prior}, we conduct experiments to verify that when the oracle noise transition is applied, how is the performance of current transition-based models and summary them as follows. From the results, we can see the method under the oracle transition achieves the best performance in most cases among these transition-based methods, except some high-noise scenarios in the symmetric noise. Regarding the counter cases, we would like to clarify that having the oracle transition does not mean the best absolutely (especially when we also consider the cross-model-training baselines like DivideMix). This is because the noise model is one of many factors in label-noise learning, and others like the accuracy of inferred labels, the model capacity and the learning mechanism are also important to the final performance. In this work, we explore a better noise-transition estimation, and also extend our method as a general plug-in to more powerful paradigms (e.g., cross-model training strategy) and demonstrate its efficiency by experiments. 
}


\ifCLASSOPTIONcaptionsoff
  \newpage
\fi



\bibliographystyle{IEEEtran}
\bibliography{references}
%



%

\begin{IEEEbiography}[{\includegraphics[width=1in,height=1.25in,clip]{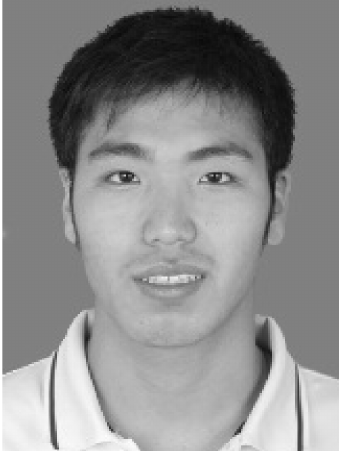}}]{Jiangchao Yao} is an Assistant Professor of Shanghai Jiao Tong University, Shnaghai China. He
received the B.S. degree in information
engineering from South China University of Technology,
Guangzhou, China, in 2013. He got a dual Ph.D. degree under the supervision of Ya Zhang in Shanghai
Jiao Tong University and Ivor W. Tsang in University of Technology Sydney.
His research interests include deep representation learning and robust machine learning.
\end{IEEEbiography}

\begin{IEEEbiography}[{\includegraphics[width = 1\textwidth,height = 0.12\textheight]{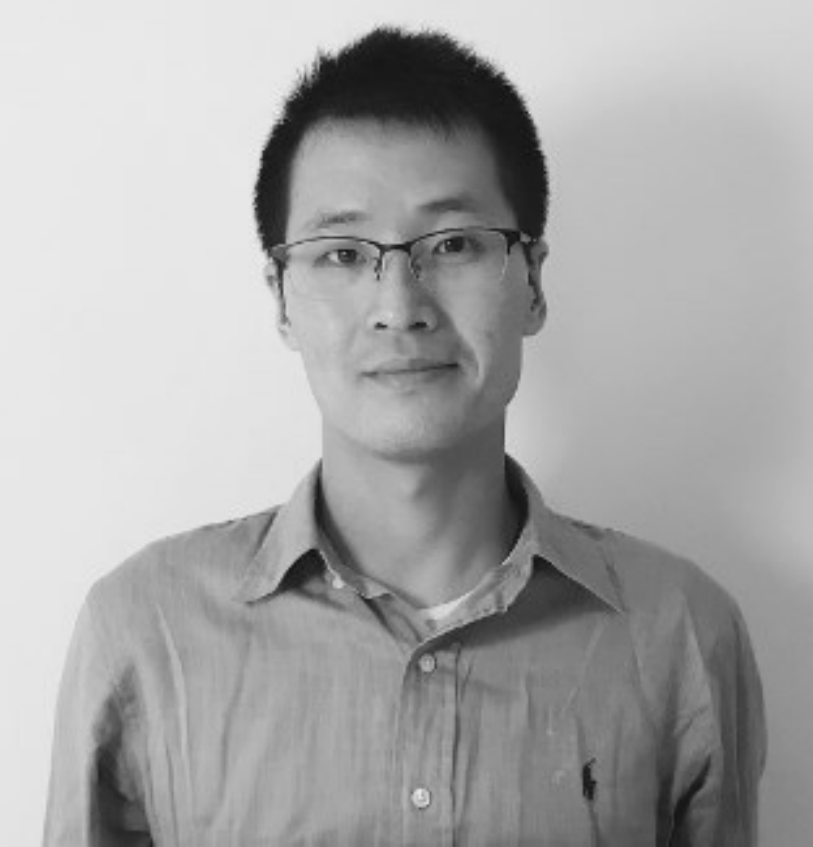}}]{Bo Han} is currently an Assistant Professor in Machine Learning and a Director of Trustworthy Machine Learning and Reasoning Group at Hong Kong Baptist University, and a BAIHO Visiting Scientist at RIKEN Center for Advanced Intelligence Project (RIKEN AIP). He was a Visiting Faculty Researcher at Microsoft Research (2022) and a Postdoc Fellow at RIKEN AIP (2019-2020). He received his Ph.D. degree in Computer Science from University of Technology Sydney (2015-2019). During 2018-2019, he was a Research Intern with the AI Residency Program at RIKEN AIP. He has co-authored a machine learning monograph, including Machine Learning with Noisy Labels (MIT Press). He has served as area chairs of NeurIPS, ICML, ICLR and UAI, and senior program committees of KDD, AAAI and IJCAI. He has also served as action (associate) editors of Transactions on Machine Learning Research and IEEE Transactions on Neural Networks and Learning Systems, and editorial board members of Journal of Machine Learning Research and Machine Learning Journal.
\end{IEEEbiography}

\begin{IEEEbiography}[{\includegraphics[width=1in,height=1.25in,clip]{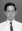}}]{Zhihan Zhou}
received the B.S. degree in information
engineering from Shanghai Jiao Tong University, Shanghai, China, in 2020. He is currently working toward the Ph.D. degree under the supervision of Ya Zhang in Shanghai
Jiao Tong University.
His research interests include machine learning and deep learning.
\end{IEEEbiography}

\begin{IEEEbiography}[{\includegraphics[width=1in,height=1.25in,clip,keepaspectratio]{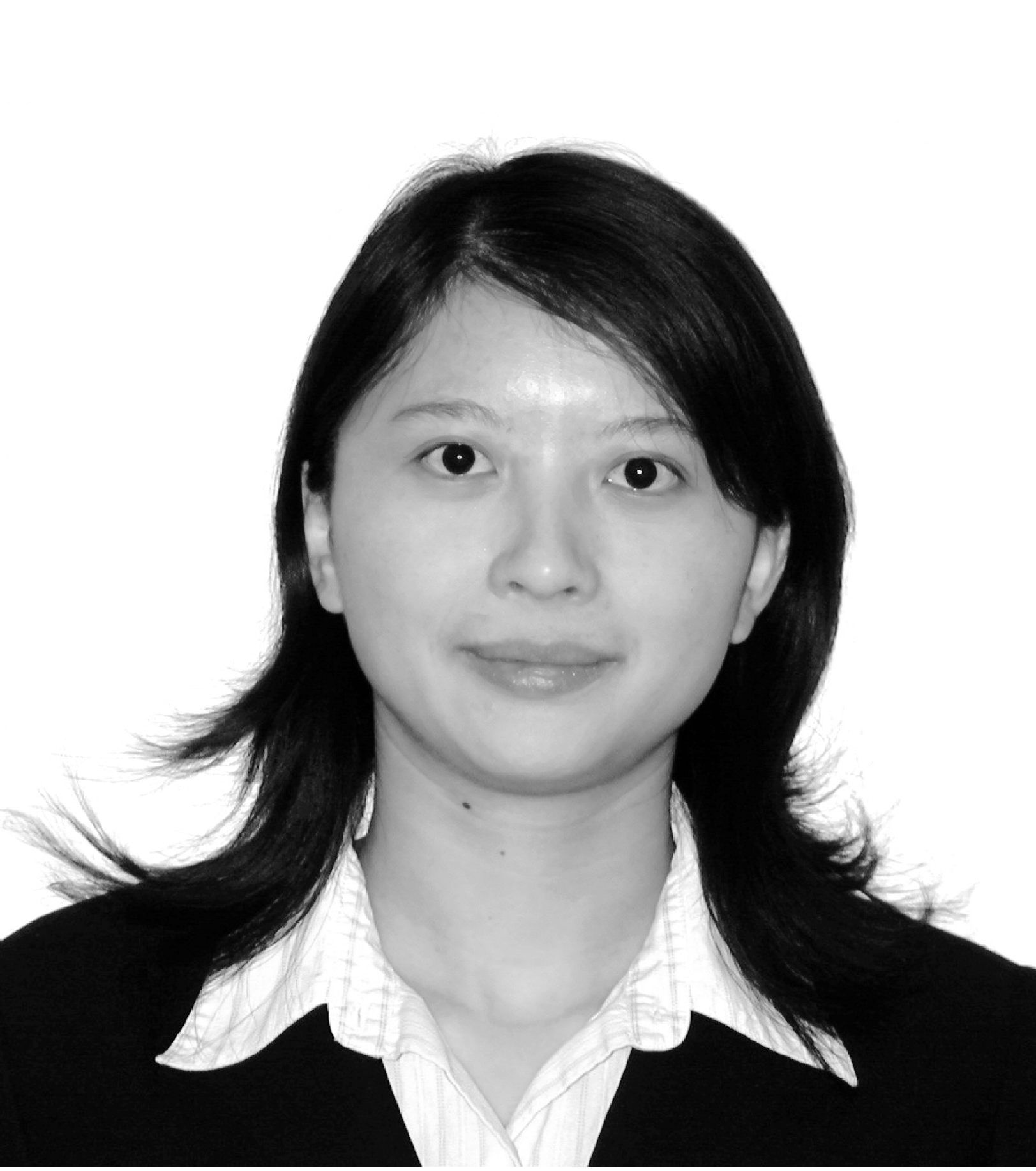}}]{Ya Zhang}
is currently a Professor in Cooperative Medianet Innovation Center, Shanghai Jiao Tong University. Her research interest is mainly on machine learning and data mining with applications to multimedia information retrieval, social network analysis, and intelligent information system. Dr. Zhang holds a PhD degree in Information Sciences and Technology from Pennsylvania State University and a Bachelor's degree from Tsinghua University in China. Before joining Shanghai Jiao Tong University, Dr. Zhang was a research manager at Yahoo! Labs, Sunnyvale, CA, USA, where she leaded a R$\&$D team of researchers with strong background in data mining and machine learning to improve the web search quality of Yahoo international markets. Prior to joining Yahoo, Dr. Zhang was an assistant professor at University of Kansas with research focus on machine learning applications in bioinformatics and information retrieval. Dr. Zhang published more than 70 refereed papers in prestigious international conferences and journals including TPAMI, TIP, TNNLS, ICDM, CVPR, ICCV, ECCV, and ECML. She currently holds 5 US patents and 4 China patents, and has 9 patents pending in the areas of multimedia analysis. She is appointed as the Chief Expert for the project 'Research of Key Technologies and Demonstration for Digital Media Self-organizing' under the 863 program by Ministry of science and technology of China.
\end{IEEEbiography}

\begin{IEEEbiography}[{\includegraphics[width=1in,height=1.25in,clip,keepaspectratio]{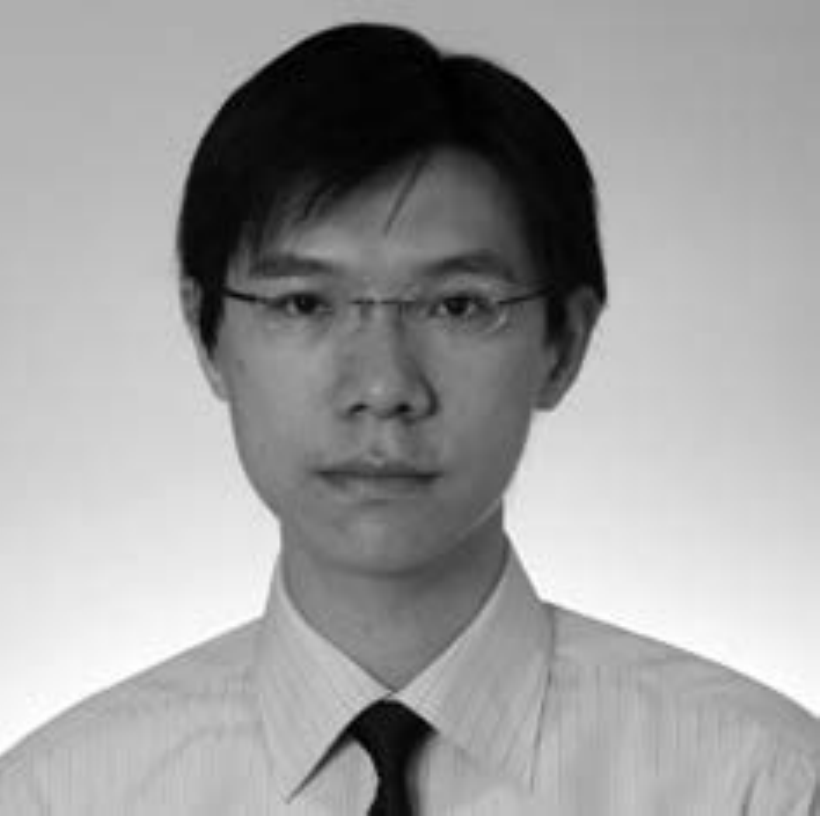}}]{Ivor W. Tsang}
is a Professor of Artificial Intelligence, at University of Technology Sydney (UTS). He is also the Research Director of the UTS Flagship Research Centre for Artificial Intelligence (CAI). His research focuses on transfer learning, feature selection, big data analytics for data with extremely high dimensions in features, samples and labels, and their applications to computer vision and pattern recognition. He has more than 190 research papers published in top-tier journal and conference papers. According to Google Scholar, he has more than 10,000 citations and his H-index is 56. In 2009, Prof Tsang was conferred the 2008 Natural Science Award (Class II) by Ministry of Education, China, which recognized his contributions to kernel methods. In 2013, Prof Tsang received his prestigious Australian Research Council Future Fellowship for his research regarding Machine Learning on Big Data. In 2019, he received the International Consortium of Chinese Mathematicians Best Paper Award in recognition of  his work "Towards ultrahigh dimensional feature selection for big data", published in Journal of Machine Learning Research. 
In addition, he had received the prestigious IEEE Transactions on Neural Networks Outstanding 2004 Paper Award in 2007, the 2014 IEEE Transactions on Multimedia Prize Paper Award, and a number of best paper awards and honors from reputable international conferences, including the Best Student Paper Award at CVPR 2010. 
\end{IEEEbiography}

\end{document}